\documentclass[lettersize,journal]{IEEEtran}
\usepackage{amsmath,amsfonts}
\usepackage{algorithmic}
\usepackage{algorithm}
\usepackage{array}
\usepackage[caption=false,font=normalsize,labelfont=sf,textfont=sf]{subfig}
\usepackage{textcomp}
\usepackage{stfloats}
\usepackage{url}
\usepackage{verbatim}
\usepackage{graphicx}
\usepackage{cite}
\usepackage{multirow}
\usepackage{amssymb}
\usepackage{accents}
\usepackage[table]{xcolor}
\usepackage{graphics}
\usepackage{float} 

\newcommand{\itunderT}[2]{\underline{\textit{#1}} \pm \underline{\textit{#2}}} 
\newcommand{\boldT}[2]{{\textbf{#1}} \pm {\textbf{#2}}} 

\begin{document}

\title{No-Clean-Reference Image Super-Resolution: Application to Electron Microscopy}

\author{Mohammad Khateri,~\IEEEmembership{ Student Member,~IEEE}, Morteza Ghahremani,~\IEEEmembership{Member,~IEEE}, Alejandra Sierra, \\ and Jussi Tohka

\thanks{This work was supported in part by the Research Council of Finland (\#323385), Jane and Aatos Erkko Foundation, and Doctoral Programme in Molecular Medicine at the University of Eastern Finland.}
\thanks{Mohammad Khateri, Morteza Ghahremani, Alejandra Sierra, and Jussi Tohka are with A. I. Virtanen Institute for Molecular Sciences, Faculty of Health Sciences, University of Eastern Finland, Finland (e-mail:  mohammad.khateri@uef.fi; morteza.ghahremani@uef.fi; alejandra.sierralopez@uef.fi; jussi.tohka@uef.fi).  Morteza Ghahremani is also with Artificial Intelligence in Medical Imaging, Department of Radiology, Technical University of Munich, Germany (e-mail: morteza.ghahremani@tum.de). } }

\markboth{IEEE Transactions Format}%
{Shell \MakeLowercase{\textit{et al.}}: A Sample Article Using IEEEtran.cls for IEEE Journals}


\maketitle
\begin{abstract}
The inability to acquire clean high-resolution (HR) electron microscopy (EM) images over a large brain tissue volume hampers many neuroscience studies. To address this challenge, we propose a deep-learning-based image super-resolution (SR) approach to computationally reconstruct clean HR 3D-EM  with a large field of view (FoV) from noisy low-resolution (LR) acquisition. Our contributions are I) Investigating training with no-clean references; II) Introducing a novel network architecture, named EMSR, for enhancing the resolution of LR EM images while reducing inherent noise. The EMSR leverages distinctive features in brain EM images—repetitive textural and geometrical patterns amidst less informative backgrounds— via multi-scale edge-attention and self-attention mechanisms to emphasize edge features over the background; and, III) Comparing different training strategies including using acquired LR and HR image pairs, i.e., real pairs with no-clean references contaminated with real corruptions, the pairs of synthetic LR and acquired HR, as well as acquired LR and denoised HR pairs. Experiments with nine brain datasets showed that training with real pairs can produce high-quality super-resolved results, demonstrating the feasibility of training with non-clean references. Additionally, comparable results were observed, both visually and numerically, when employing denoised and noisy references for training. Moreover, utilizing the network trained with synthetically generated LR images from HR counterparts proved effective in yielding satisfactory SR results, even in certain cases, outperforming training with real pairs. The proposed SR network was compared quantitatively and qualitatively with several established SR techniques, showcasing either the superiority or competitiveness of the proposed method in recovering fine details while mitigating noise.
\end{abstract}


\section{Introduction}
\IEEEPARstart{T}{hree-dimensional} electron microscopy (3D-EM) is an essential technique to investigate brain tissue ultrastructures as it allows for their 3D visualization at nanometer resolution \cite{hildebrand2017whole,zheng2018complete}. Studying brain tissue ultrastructures requires high-resolution (HR) images over a large field of view (FoV) of the brain tissue. However, since imaging at higher resolutions demands denser sampling, it takes more time, proportionally increasing imaging cost and potential sample damage. Moreover, HR imaging over a large FoV is not feasible under realistic imaging constraints, demanding a trade-off between imaging resolution and FoV. The higher resolution, the smaller FoV \cite{varsano2022electron}. Furthermore, imperfect components of imaging systems introduce noise into the images \cite{roels2018overview}. The mentioned limitations collectively prevent acquiring clean HR EM images over a large FoV of brain tissue, impeding subsequent brain ultrastructure analysis and visualization.

A practical approach to mitigate such limitations to provide clean HR EM images over a large tissue volume includes the following steps: I) low-resolution (LR) imaging of brain samples over a large FoV of interest, II) HR imaging over a small but representative portion of the same samples covered by LR FoV, and III) utilizing image super-resolution (SR) technique to computationally reconstruct high-quality HR 3D-EM images from the LR 3D-EM images 
of brain tissue, which is typically contaminated with noise, artifacts, and distortions.

SR is a low-level vision task that can serve as an integral preprocessing step for many image analyses in neuroscience\cite{mikula2015high,imbrosci2022automated, funke2018large}. It aims to recover the latent clean HR image $x$ from a degraded LR observation $y$:
\begin{equation}
\label{eq_1}
y = \mathcal{D}_{\delta}(x),
\end{equation}
where $\mathcal{D}_{\delta}(\cdot)$ is the degradation function parameterized by $\delta$, which is non-invertible, making SR an ill-posed inverse problem. $\mathcal{D}_{\delta}(\cdot)$ includes convolution operator $\circledast$ with the blur kernel $\kappa$, s-fold under-sampling operator $\downarrow_s$, and noise $n$ ($\delta = \{\kappa, \downarrow_s, n\}$) \cite{liu2022blind}. In practice, $\delta$ is unknown and we only have the LR observation. 

SR methods can be categorized into two groups: model-based and learning-based methods. Model-based SR methods approximate the degradation function in (\ref{eq_1}) as a combination of several operations. Assuming that the blurring kernel and under-sampling operator are known and noise is additive:
\begin{equation} \label{eq_2}
\mathcal{D}_{\delta}(x) = (x \circledast \kappa) \downarrow_s + n 
\end{equation}
An estimate  $x^{*}$ of an HR image can then be obtained by Maximum A Posteriori (MAP) formulation as:
\begin{equation} \label{eq_3}
x^{*} = \arg\min_{x}\{ {\Vert y - (x \circledast \kappa) \downarrow_s \Vert}^q_p + \lambda\mathcal{R}(x)\}
\end{equation}
The first term is likelihood computed as the $\ell_p$-norm distance between the observation $y$ and degraded latent image $x$, where $0 < p,q\leq 2$ are determined by noise distribution \cite{ren2019simultaneous, bouman2022foundations, meng2013robust,cao2016robust}. $\mathcal{R(\cdot)}$ is the regularization term, also known as prior term, penalizing unknown latent image $x$ upon our prior knowledge of data. The parameter $\lambda$ defines the trade-off between likelihood and prior terms. To reduce the ill-posedness of SR problems, many regularization terms have been developed \cite{wang2020deep, liu2022blind}, each with specific pros and cons. Notably, contributions from total variation \cite{rudin1992nonlinear}, self-similarity \cite{glasner2009super}, low-rankness \cite{cao2016robust}, and sparse representation \cite{zha2022low} have played a significant role in improving SR performance, among others. Crafted priors enhance SR but have limited performance compared to data-driven methods \cite{liu2022blind}. Effective SR models involve optimizing multiple priors, which is time- and memory-consuming, and require tuning trade-off parameters. Additionally, SR models are specific to certain degradation settings, necessitating separate models for each degradation. Mismatched LR images with different degradations may result in severe artifacts due to domain gaps \cite{chen2022real}.

Learning-based SR methods learn a mapping between LR and HR image spaces, which is then used to restore the HR image from the given LR input image. Early work, pioneered by \cite{freeman2000markov}, restored HR images by capturing the co-occurrence prior between LR and HR image patches. Numerous patch-based methods have been introduced relying on manifold learning \cite{lu2013image}, filter learning \cite{romano2016raisr}, regression \cite{yang2013fast}, and sparse representation \cite{yang2010image}. Deep neural network (DNN)-based SR methods have demonstrated remarkable performance 
\cite{wang2020deep}. DNNs with end-to-end training avoid the need for explicit design of priors or degradations. Instead, priors and degradations are encapsulated in the training datasets.
The commonly used DNN architectures include convolutional neural networks (CNNs) \cite{dong2014learning, song2020super}, generative adversarial networks (GANs) \cite{zhang2021designing, sui2022scan, wang2021real}, vision transformers (ViTs) \cite{lu2022transformer,liang2021swinir}, and denoising diffusion probabilistic models (DDPMs) \cite{saharia2022image, mei2023conditional}. In this realm, many works in computer vision and biomedical imaging define a specific degradation function to synthesize LR images from HR counterparts to generate the training data\cite{liu2022blind}.
Several studies were also conducted to incorporate the interpretability of model-based methods into end-to-end learning, e.g., deep unfolding \cite{zhang2020deep, ma2021deep, karl2023foundations}, Plug-and-Play (PnP)\cite{chan2016plug, zhang2021plug, shoushtari2023prior, abu2022image}, and deep equilibrium learning\cite{zou2023deep, gilton2021deep}. Although most of these degradation-oriented SR approaches lead to satisfactory results on benchmark datasets, they fail to restore high-quality images when it comes to real-world applications \cite{chen2022real}, such as brain EM images, which is the focus of this study. 

The computational approaches in super-resolution of EM have been studied in health and material sciences \cite{tsiper2017sparsity, sreehari2017multi, gao2020deep, fang2021deep}. As a pioneer, \cite{sreehari2017multi} proposed a material-specific PnP approach to super-resolve LR EM. Their method was based on the MAP formulation, where the likelihood term was based on a linear degradation model and the prior term was a library-based non-local means (LB-NLM) designed on HR EM images acquired within small FoV. The presence of HR edges and textures corresponding to the LR input image in the designed library yielded super-resolved results with fine details. To reduce the computational expenses and improve generalization, authors in \cite{reid2022multi} replaced the LB-NLM denoiser with the off-the-shelf Gaussian denoiser, leading to the version of PnP typically used in biomedical 
applications. However, both methods \cite{sreehari2017multi,reid2022multi} are essentially model-based, computationally cumbersome, and limited to degradation models. Experiments in both studies were conducted on the EM datasets acquired from nano-material with simple textural information, which sparsely recurred throughout the image. By leveraging the unique characteristics of such images, authors in \cite{qian2020effective} devised a patch-based strategy on acquired pairs of LR and HR EM images in the training of LB-NLM, resulting in better performance than the original LB-NLM method but inferior to DNN-based methods. Authors in \cite{fang2021deep} introduced a DNN-based SR,  named point scanning super-resolution (PSSR), for EM brain images. They proposed a degradation operator, i.e., crappifier, to synthesize LR images from acquired HR counterparts, where the crappifier included additive Gaussian noise followed by a down-sampling operator. Using synthetic pairs of LR and HR EM images, they trained a UNet-based residual neural network. The performance of the method was then compared only with the bilinear interpolation. Although synthetizing pairs of LR and HR EM images can reduce imaging costs, it can increase the domain gap between the input LR EM and the trained SR model. 

DNN-based SR method can implicitly learn EM degradations if trained with acquired and matched pairs of LR and HR images. However, many challenges impede designing DNN-based frameworks with such data. First, EM images inherently contain noise and artifacts derived from the microscope, sample, and experimental settings. Hence, there is no clean EM image to be used as the reference for training the network. Further, networks pre-trained on natural images cannot restore high-quality brain EM images due to the considerable difference in the physics behind photography and EM as well as content dissimilarity between natural and brain EM images. Hence, deploying and designing SR methods for EM images demands specific considerations. In this work, we will illustrate and address the mentioned challenges of the SR of EM images. Our key contributions are as follows:
\begin{itemize}
    \item Investigating training using no-clean references for $\ell_2$ and $\ell_1$ loss functions. 
    \item Introducing a DL-based image SR framework for EM, named EMSR, equipped with edge-attention and self-attention mechanisms for enhanced edge recovery. Sharing the network's modules between the original noisy LR EM image and its noisier version makes it noise-robust.
    \item Comparing various training strategies focusing on EM images, including training from pairs of physically acquired LR and HR, synthetically generated LR and HR, as well as LR and denoised HR EM images. 
\end{itemize}
The remainder of this article is organized as follows: Section II describes the proposed image super-resolution method, Section III delves into experimental results, and finally, Section IV concludes the article.

\section{Proposed Method}

\begin{figure*}[!t]
\centering\vspace{-3mm}
\includegraphics[width=0.78\textwidth,height=0.32\textwidth]{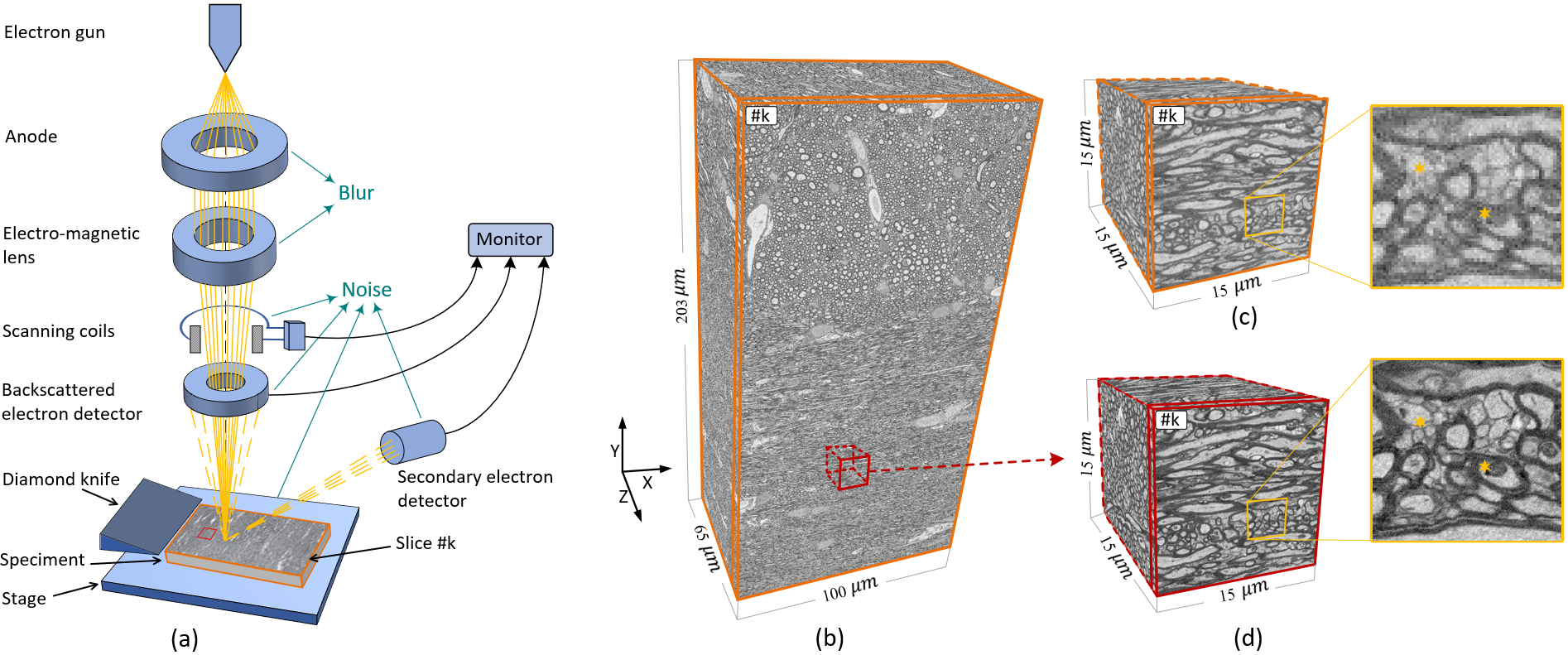}
\caption{Schematic diagram of serial block-face scanning electron microscopy and imaging. a) The electron gun generates streams of electrons that are focused and raster scanned across the sample's surface (solid yellow lines). The interaction of these focused electrons with the sample results in the ejection of electron streams (dashed yellow lines), which are collected by detectors to form a 2D image of $k$-\textit{th} slice (labeled as $\small{\#}k$). Note that the region of interest from the sample is imaged at LR with a large FoV (marked in orange), while at HR, the FoV is smaller (marked in red). After imaging a slice, the diamond knife is used to cut the sample to a specific thickness, determining the resolution in the $z$ direction and exposing the subsequent block-face for imaging. Imperfections in the imaging device components can introduce blurring and noise in the resultant images (solid green arrows). b) Stack of 2D imaged slices constitutes the 3D-EM dataset. c) LR 3D-EM corresponding to d) HR 3D-EM from small FoV. The zoomed-in area from (c) and (d) demonstrates the superior quality of the HR image in terms of contrast and resolution, see asterisks.}\label{Fig_schematic_diagram}
\end{figure*}

Supervised training of a network requires numerous pairs of corrupted LR and corresponding clean reference images. However, brain EM images inevitably include different types of noise, artifacts, and distortions, caused by the imaging system, and experimental settings. Therefore, clean EM images that serve as references are unavailable. Here, we investigate training a neural network for EM SR using physically acquired pairs of LR and HR EM images contaminated with real noise-like corruptions.

\subsection{Electron Microscopy Super-Resolution}
In serial block-face scanning electron microscopy (SBEM), a focused high-energy electron beam scans the sample surface, resulting in the acquisition of a 2D image in $xy$-plane. The diamond knife subsequently removes the top layer of the sample to a specific thickness in $z$ direction, revealing the next block-face for imaging. The repetition of this process generates a series of 2D images that are stacked to form a 3D volume image, as illustrated in Fig. \ref{Fig_schematic_diagram}.

The observed block-face $y \in \mathbb{R}^{m \times m}$ is affected by underlying microscope degradation $\mathcal{D}_{\delta^{\prime}}(\cdot): \mathbb{R}^{M \times M} \rightarrow \mathbb{R}^{m \times m}$ parameterized by $\delta^{\prime}$, $y = \mathcal{D}_{\delta^{\prime}}(x)$; where, $ x \in \mathbb{R}^{M \times M}$ denotes the latent image that we aim to restore, $ M = \tau m$ wherein  $\tau$ is the resolution ratio between HR and LR images, i.e., super-sampling ratio. Theoretically, the SR process is to recover unknown $x$ via  $\mathcal{D}_{\delta^{\prime}}^{-1}(y)$, demanding finding degradation inversion $\mathcal{D}_{\delta^{\prime}}^{-1}(\cdot) : \mathbb{R}^{m \times m} \rightarrow \mathbb{R}^{M \times M}$. If such a mapping exists, we can obtain HR observations through LR imaging, practically accelerating imaging by a factor $ \tau ^2$.
Microscope degradation parameters, $\delta^{\prime}$, can arise from various sources \cite{roels2018overview, titze2013techniques, de2015influence}. These sources include electronic device components such as wires and coils, which produce thermal and electromagnetic interference that is modeled as Gaussian noise. The detector's electron-counting error introduces signal-dependent noise in EM images, which is modeled as Poisson noise. Line-by-line pixel scanning in SBEM can lead to correlated noise. Imperfect electromagnetic lenses and anodes cause blurred observations due to suboptimal focusing of the electron beam. The high-energy electron beam introduces electron charge and causes absorption-based heating. Cutting the sample with a diamond knife can introduce specific artifacts and distortions. Additionally, mechanical disturbances from the environment and microscope can introduce mechanical noise, further exacerbating image degradation.

Hence, $\mathcal{D}_{\delta^{\prime}}(\cdot)$ cannot be well parameterized by simplified assumptions such as block-averaging neighbor pixels for the under-sampling operator \cite{reid2022multi}.
Implicit modeling of degradation function can be realized through training a neural network by acquired pairs of LR and HR EM images.
\begin{figure*}[!t]
\centering
\includegraphics[width=0.65\textwidth,height=0.34\textwidth]{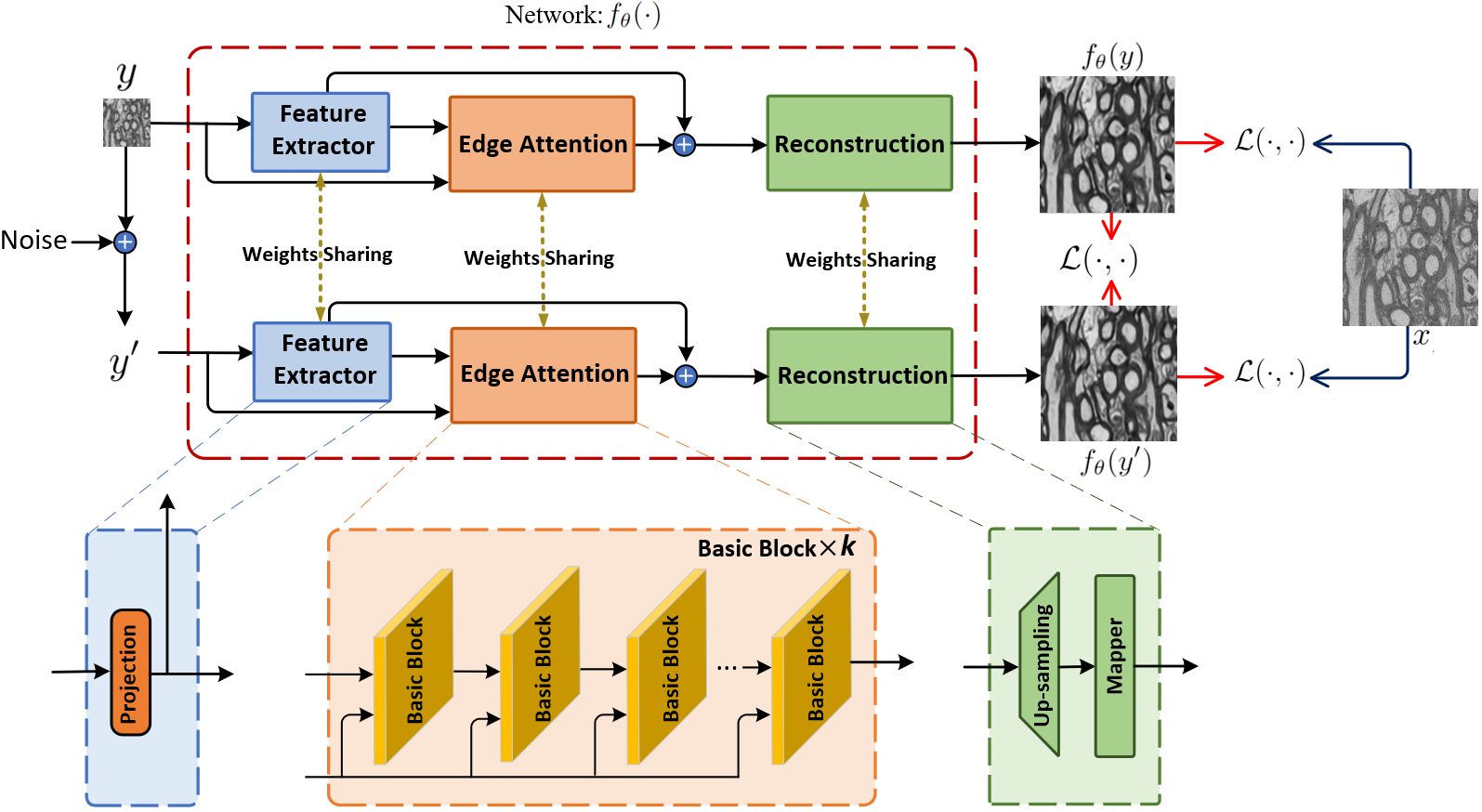}\vspace{-2mm}
\caption{Overview of the proposed image super-resolution network for training with pairs of corrupted images. The network includes the feature extractor, edge attention, and reconstruction modules, which are shared between the original noisy LR EM image $y$ and its noisier version $y^{\prime}$. The network is encouraged to generate two outputs, $f_{\theta}(y)$ and $f_{\theta}(y^{\prime})$, that are consistent with the noisy reference image $x$. The output from the original image $f_{\theta}(y)$ serves as the reference for the noisier-noisy input, establishing a noise-robust framework in a self-supervised approach.}
\label{fig_network_overview}
\end{figure*}

\subsection{Training without Clean Reference}
Training with no-clean references has been studied in several image restoration tasks, including denoising, magnetic resonance image reconstruction, and text removal \cite{pmlr-v80-lehtinen18a, moran2020noisier2noise, calvarons2021improved}. Here, our focus is on investigating such a training approach for commonly used restoration loss functions, i.e., $\ell_2$ and $\ell_1$, and discussing the corruption levels at which this training remains feasible for EM SR.

Supervised training of a network $f_{\theta}(\cdot)$ for SR requires numerous pairs of degraded LR, $y$, and clean reference, $x$. The network's parameters $\theta$ are obtained by optimizing the following empirical loss function:
\begin{equation}\label{eq_6}
\hat{\theta} =  \arg\min_{\theta}{ \mathbb{E}_{(x,y)} [\mathcal{L}(f_{\theta}(y),x)] }
\end{equation}
By applying the conditional expectation rule for dependent random variables $y$ and $x$, we can reformulate (\ref{eq_6}) as follows:

\begin{multline}\label{eq_7}
\hspace{16mm} \hat{\theta} =  \arg\min_{\theta}{ \mathbb{E}_{y}[ \underbrace{\mathbb{E}_{x|y}[\mathcal{L}(f_{\theta}(y),x)]}_{\text{reference-dependent}}] } \hspace{10mm}
\end{multline}
The equation above implies that the network parameters can be optimized separately with respect to $y$ and $x$ over the loss function $\mathcal{L(\cdot,\cdot)}$. Let $\hat{x} = x + n$, where $n$ is an i.i.d.  additive noise with mean $\mu$ and variance $\sigma_n^2I$, where $I \in \mathbb{R}^{d\times d}$ is an identity matrix with $d=M^2$.

In the case loss function is $\ell_2$, we can derive equality that links the solutions of the reference-dependent component in (\ref{eq_7}) for $x$ and $\hat{x}$ as follows (see Appendix I.A):
\begin{equation} \label{eq_const_l2}
\begin{split}
&\mathbb{E}_{\hat{x}|y}[\|f_{\theta}(y)-\hat{x}\|_{2}^{2}]=\\
&\mathbb{E}_{x|y}[\|f_{\theta}(y)-x\|_{2}^{2}] -2\mu^T\mathbb{E}_{x|y}[f_{\theta}(y)-x] + d\sigma_{n}^2 + ||\mu||^2
\end{split}
\end{equation}
The equation above states that when $\mu$ is close to zero ($\mathbb{E}[n] \approx 0$), the second term on the right-hand side of the equation becomes negligible, i.e., $2\mu^T\mathbb{E}_{x|y}[f_{\theta}(y)-x] \to 0$. Additionally, the third term $ \sigma_{n}^2$, which is noise variance, and the fourth term, which is noise mean, are independent of $y$ and have no effect on the total optimization problem. Therefore, if we substitute the clean image $x$ with a random variable $\hat{x}$ that satisfies $\mathbb{E}[x] \approx \mathbb{E}[\hat{x}]$, the network's parameters will remain close to the optimal. This enables us to replace the clean reference $x$ with its corrupted version $\hat{x}$, provided their expectation values are sufficiently close, which can be accompanied by the practical assumption that noise should not significantly alter the overall variability and structure of the original image, i.e., $\sigma_{\hat{x}}^2 \approx \sigma_x^2$. 

In the case of $\ell_1$ loss, we can establish the relationship between the solutions of the reference-dependent part in (\ref{eq_7}) for both $x$ and $\hat{x}$ as below (see Appendix I.B):
\begin{multline}\label{eq_const_l1}
\Big| \mathbb{E}_{\hat{x}|y}[\|(f_{\theta}(y)-\hat{x})\|_{1}] - \mathbb{E}_{x|y}[\|f_{\theta}(y)-x\|_{1}] \Big|  \\
\leq \frac{| -2\mu^T\mathbb{E}_{x|y}[f_{\theta}(y)-x] + d\sigma_{n}^2 + ||\mu||^2 |}{g(y,x,\hat{x})},
\end{multline}
where $g(y,x,\hat{x}) = \frac{ \sqrt{\mathbb{E}_{\hat{x}|y}[\|f_{\theta}(y)-\hat{x}\|_{2}^{2}]} + \sqrt{\mathbb{E}_{x|y}[\|f_{\theta}(y)-x\|_{2}^{2}]} }{\sqrt{d}} $. The inequality above suggests that the difference between the reference-dependent solutions for $\hat{x}$ and $x$ is bounded by a function of $\mu$ and $\sigma_n^2$. When $\mu$ is small, it significantly reduces the dependence on $y$ and tightens the upper bound, which becomes primarily dependent on $y$ through $\sigma_n^2$. This implies that weak noise reduces the reliance on $y$ and indicates that it will not significantly alter the overall optimization problem (\ref{eq_7}).  In other words, the network's parameters will remain near optimal even if we replace clean image $x$ with its noisy version $\hat{x}$, as long as $\mathbb{E}[x] \approx \mathbb{E}[\hat{x}]$, and overall structure of the clean image is not significantly altered by noise, $\sigma_{\hat{x}}^2 \approx \sigma_x^2$.

These observations hold the promise that the network can be trained under real-world scenarios where the reference is contaminated with weak noise-like corruptions. Here, we aim to determine the rough acceptable level of these corruptions in brain EM imaging--which was discussed in Section II. (B), upon noise statistics $\mu$ and $\sigma^2$ that overshadow training with no-clean references. Suppose we can decompose $\hat{x}$ into clean $x$ and noise-like corruption component $n$, $\hat{x}= x_{clean} + n$. We can then establish the following relationships:
\begin{subequations}
\label{eq_temp}
\begin{align}
    \label{eq_temp:I}
     \mathbb{E}[\hat{x}] &= \mathbb{E}[x_{clean}] +\mathbb{E}[n], \\
    \label{eq_temp:II}
    \sigma_{\hat{x}}^2 &= \sigma_{x_{clean}}^2 + \sigma_n^2
\end{align}
\end{subequations}
The inequalities (\ref{eq_10:I}) and (\ref{eq_10:II}) hold $\mathbb{E}[x_{clean}] \approx \mathbb{E}[\hat{x}]$ and $\sigma_{x_{clean}}^2 \approx \sigma_{\hat{x}}^2$, which are requirements for training using pairs of corrupted images using $\ell_1$ and $\ell_2$ loss functions, and guarantee that the content of the underlying image is much stronger than corruptions:
\begin{subequations}
\label{eq_10}
\begin{align}
    \label{eq_10:I}
    \mathbb{E}[x_{clean}] &  \gg  \mathbb{E}[n] , \\
    \label{eq_10:II}
    \sigma_{x_{clean}}^2 &  \gg \sigma_n^2
\end{align}
\end{subequations}
The level of corruptions in EM is mostly much lower than image content information, satisfying (\ref{eq_10}), allowing for training network $f_{\theta}(\cdot)$ from pairs of corrupted images. 
It is  worth mentioning that rare image slices may exhibit levels of corruption inconsistent with constraints stated in (\ref{eq_10}). These corruptions act as anomalies that the network is unable to learn.

\begin{figure*}[!t]
\centering
\includegraphics[width=0.7\textwidth,height=0.30\textwidth]{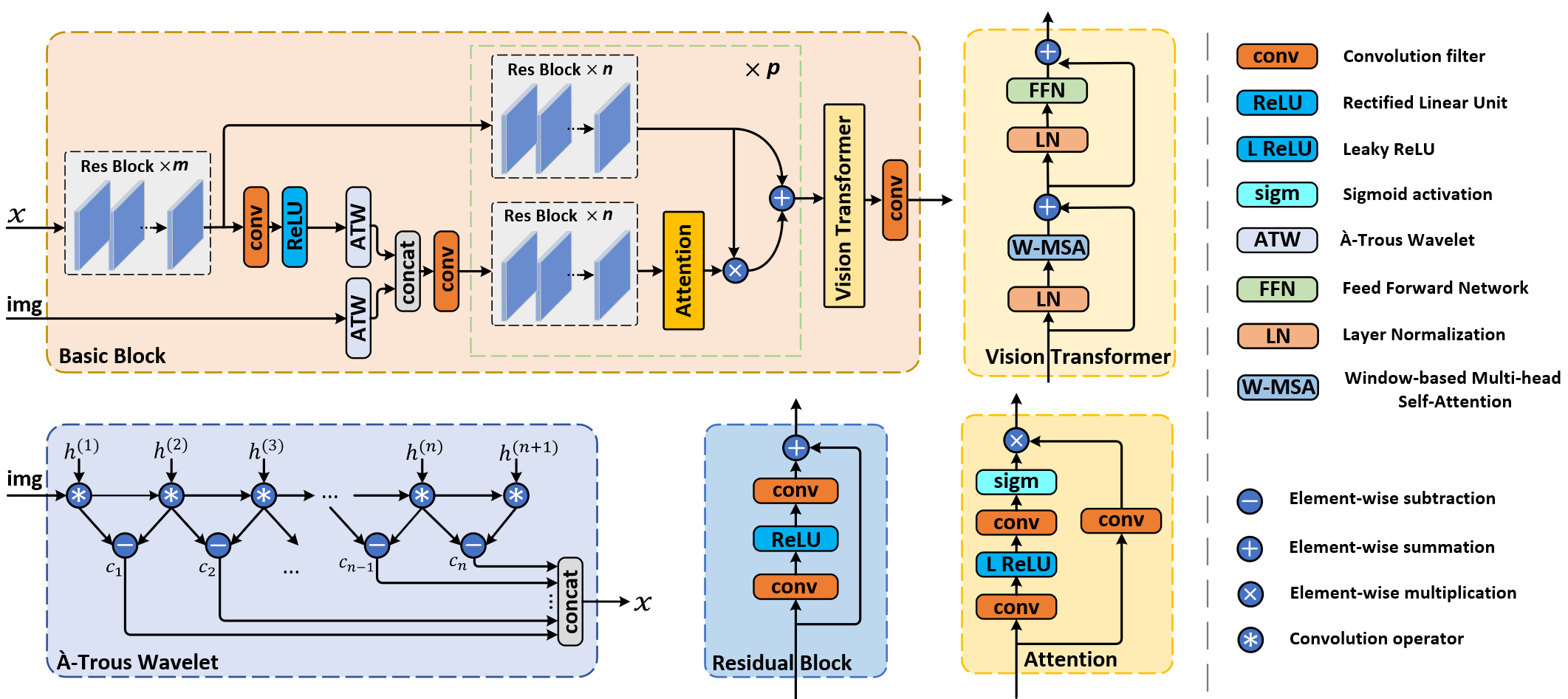} \vspace{-3mm}
\caption{Modules embedded in the proposed network: Basic Block, Residual Block, À-Trous Wavelet, Attention Block, and Vision Transformer Block.}\label{fig_modules_network}\end{figure*}

\subsection{Network Architecture}
The proposed SR network, which is designed for training using pairs of corrupted LR and HR EM images, is depicted in figure \ref{fig_network_overview}. It consists of three key modules: feature extractor, edge attention, and reconstruction. These modules are shared between the given LR image and its noisier version.

\subsubsection{Feature Extractor} The feature extractor ($\mathcal{H}_{FE}$) is employed to extract shallow ($X_{SF}$) features from the given LR image $\mathbf{y} \in \mathbb{R}^{W \times H \times C}$. It includes a projection, which is a $3 \times 3$ convolutional filter. The extraction process is formulated by:
\begin{equation}
\begin{split}
X_{SF} = \mathcal{H}_{FE}(y)
\end{split}
\end{equation}

\subsubsection{Edge Attention} The edge attention module ($\mathcal{H}_{EA}$)  takes $X_{SF}$ and $y$ as input, extracting deep features and combining them with edge information using multi-scale edge attention and self-attention mechanisms, yielding the generation of edge-attentioned features ($X_{EA}$). 
The calculation of the edge-attention module is summarized as:
\begin{equation} 
\begin{split}
X_{EA}=\mathcal{H}_{EA}(y, X_{SF})
\end{split}
\end{equation}
The module consists of $k$ basic blocks, as shown in figure \ref{fig_modules_network}. In each basic block, the input features pass through $m$ residual blocks with the well-studied benefits \cite{he2016deep}, then, take two parallel paths. In the upper path, the features are fed into residual blocks to produce deep features that are then enhanced using edge information. In the lower path, 
the features go through convolutional operations and ReLU activation to reconstruct the image in the LR space. The reconstructed image, along with the original LR image, is then fed to atrous wavelet (ATW) \cite{ghahremani2020ffd}, a noise-robust feature extractor, to extract multi-scale edges. The resulting multi-scale edge features are then subjected to concatenation and filtering before being inputted into the attention block. The attention block generates multi-scale attention maps specifically focused on the deep feature edges. Finally, attention maps and deep features are combined through element-wise multiplication. The resulting attentioned-features are then added to the features from the upper path, leading to the generation of multi-scale edge-attention features. Subsequently, these features are passed into a ViT block, employing a window-based multi-head self-attention mechanism to capture both local and global image dependencies within the deep multi-scale edge attention features, and finally pass through convolution layers.

\textbf{Vision Transformer (ViT):} ViTs divide a feature map into a sequence of small patches, forming local windows, and utilize self-attention mechanisms to understand the relationships among them. This capacity to comprehend diverse image dependencies is crucial for representation learning performance in low-level vision tasks such as SR. To capture both global and local image dependencies while maintaining computational efficiency, we adopt the window-based multi-head self-attention (W-MSA) method~\cite{liang2021swinir}. The attention maps generated by W-MSA are then processed through the feed-forward network (FFN). These W-MSA and FFN components are integrated into a ViT block, as illustrated in Figure \ref{fig_modules_network}, and their computations are outlined below:
\begin{equation}
\begin{split}
&{X^{\prime}} = \text{W-MSA}(\text{LN}({X})) + X,\\
&{X^{\prime\prime}} =  \text{FFN}(\text{LN}({X}^{\prime}))+ {X}^{\prime},
\end{split}
\end{equation}
where, $\text{LN}$ is layer normalization  and $X$ is the input feature map. 

In the W-MSA, the input feature map of size $C \times H \times W$ is initially divided into $N=HW/M^2$ non-overlapping local windows of size $M \times M$, resulting in local feature maps $X \in \mathbb{R}^{M^2 \times C}$. Each of these local feature maps then undergoes the standard self-attention mechanism, with the following calculation: 
\begin{equation}
Q = XP_Q,\hspace{5mm} K = XP_K,\hspace{5mm} V = XP_V,
\end{equation}
where, $P_Q, P_K, P_V \in \mathbb{R}^{C \times d_k}$ represent the query ($Q$), key ($K$), and value ($V$) projection matrices, respectively; $d_k$ is determined as $C/k$, where $k$ denotes the number of attention heads. The attention matrix is computed using the self-attention mechanism within $k$-th head of local window:
\begin{equation}
\text{Attention}({Q},{K},{V}) = \text{SoftMax}({Q}{K}^{T}/\sqrt{d_k}){V},
\end{equation}
The concatenation of all attention heads results in the multi-head self-attention (W-MSA) output. 

FFN is a multi-layer perceptron (MLP) used to introduce additional non-linearity to the model through two fully connected layers and ReLU activation.

\subsubsection{Reconstruction} Shallow features predominantly consist of low frequencies, capturing the overall structure, while the deep features encompass high frequencies corresponding to lost fine details. The long skip connection provides the reconstruction module with low frequencies and makes the training more stable. Further, it helps the edge-attention module focus on learning fine details. The element-wise summation of shallow and deep features in the LR space are fed to the reconstruction module ($\mathcal{H}_{R}$) to generate super-resolved image $x$ with enhanced resolution:
\begin{equation} 
\begin{split}
x = \mathcal{H}_{R}(X_{EA} + X_{IF})
\end{split}
\end{equation}
The reconstruction module includes an up-sampling process that enlarges the features by pixel shuffling \cite{shi2016real}. This up-sampling step is followed by a mapper module including convolution layers, which yields the super-resolved image.

\subsubsection{Weight Sharing} The aforementioned modules are shared between the given LR EM image and its noisier version, as illustrated in figure \ref{fig_network_overview}. The weight sharing encourages the network to produce consistent outputs for both the given LR image and its noisier version, establishing a noise-robust framework for training. This strategy mitigates the absence of a clean reference: The prediction generated from the given LR EM image serves as a reference for the noisier LR EM branch in a self-supervised approach.

\subsection{Loss Functions}
We employ the $\ell_p$-norm loss, $p\in\{1,2\}$, as a pixel-wise distance measure between the network's prediction $\hat{z}$ and ground truth $z$: $ \mathcal{L}_{\ell_p}(z,\hat{z}) =  {{\Vert z_i - \hat{z_i} \Vert}_p^p}$ \cite{wang2020deep, zhao2016loss}. Our loss function measures the mismatch between the two network outputs and the reference, namely $\mathcal{L}_{\ell_p}(f_{\theta}(y),x)$ and $\mathcal{L}_{\ell_p}(f_{\theta}(y^{\prime}),x)$, as well as the mismatch between two outputs, $\mathcal{L}_{\ell_p}(f_{\theta}(y),f_{\theta}(y^{\prime}))$, see Figure \ref{fig_network_overview}. The total loss is then defined as: 
\begin{equation}
\label{eq:loss_function}
 \mathcal{L}_{T} =  \lambda_1\mathcal{L}_{\ell_p}(f_{\theta}(y),x) + \lambda_2\mathcal{L}_{\ell_p}(f_{\theta}(y^{\prime}),x) + \lambda_3\mathcal{L}_{\ell_p}(f_{\theta}(y),f_{\theta}(y^{\prime})) 
\end{equation}
where $\lambda_1$, $\lambda_2$, and $\lambda_3$ are hyperparameters that govern the trade-off between components.

\begin{figure}[!t]
\centering
\includegraphics[width=0.40\textwidth,height=0.21\textwidth]{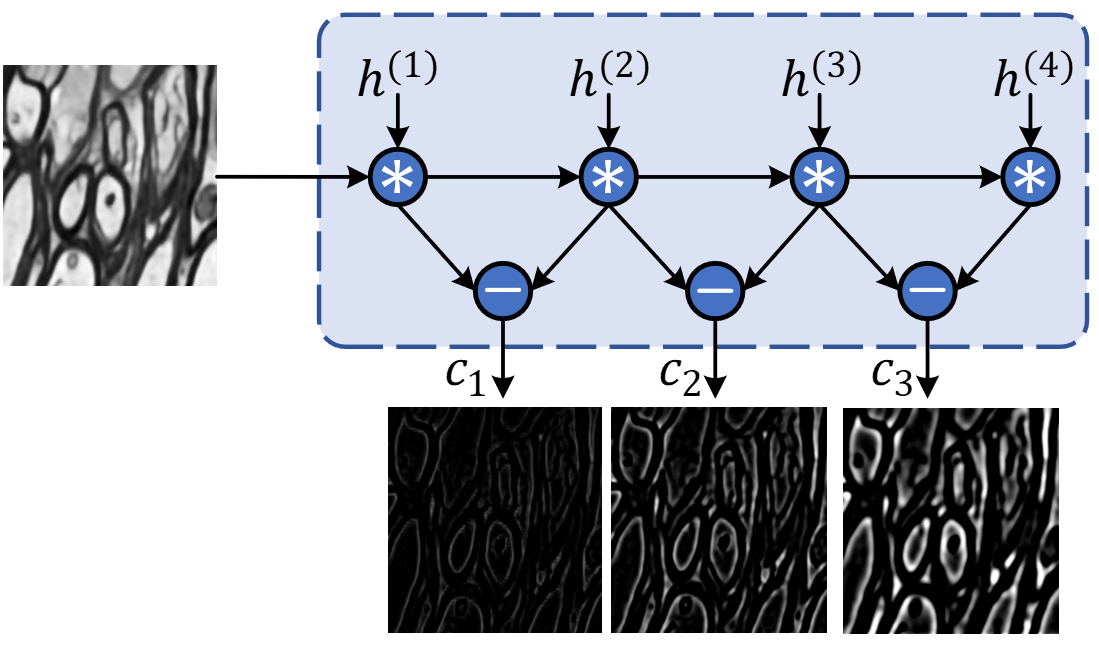} \vspace{-2.5mm}
\caption{Multi-scale edges extracted from the EM dataset using ATW, where $h$ represents the filter's kernel. The figure illustrates three edge components obtained at different scales, demonstrating the sparsity of edges and underscoring the importance of paying attention to edge details.}\label{fig_EM_edges}
\end{figure}
\vspace{-2mm}
\section{Experimental Settings and Results}
\subsection{Datasets}
We conducted experiments using nine LR and HR 3D-EM datasets acquired from the corpus callosum and cingulum regions associated with the white matter of five rat brains \cite{abdollahzadeh2021deepacson}. These datasets were acquired both ipsi- and contra-laterally. In four animals, both ipsi- and contra-lateral datasets were available, while in one animal, only ipsi-lateral data was available. Both LR and HR datasets were acquired simultaneously using the SBEM technique. The LR datasets were obtained from large tissue volumes of $200 \times 100 \times 65 \mu m^3$, with a voxel size of $50 \times 50 \times 50 nm^3$. While the HR datasets were acquired from smaller tissue volumes of $ 15 \times 15 \times 15 \mu m^3$, which were covered by the LR FoV, with a voxel size of $15 \pm 2.5 \times 15 \pm 2.5 \times 50 nm^3$. The LR and HR 3D-EM datasets totaled approximately two hundred gigabytes in size. The pairs of LR and HR from small FoV were utilized in the experiments. In terms of dimensions, the LR and HR 3D-EM pairs had size ranges respectively within $330 \pm 40 \times 330 \pm 40 \times 550 \pm 150$ and $1024 \times 1024 \times 550 \pm 150$ voxels. Animal procedures were approved by the Committee of the Provincial Government of Southern Finland, following European Community Council Directives 86/609/EEC.

\subsection{Settings}
\subsubsection{Training} Datasets were augmented by adding random zero-mean white Gaussian noise with a standard deviation of $\sigma \in [0, 5]$, applying random rotation of $\theta \in \{90^\circ, 180^\circ, 270^\circ \}$, and horizontal/vertical flipping on the input data. The noisier version of the input image was generated by adding random zero-mean white Gaussian noise with a standard deviation of $\sigma \in [0, 5]$. The Network was optimized using Adam \cite{kingma2014adam} for $200,000$  steps. The initial learning rate was set to ${10}^{-4}$ and halved every $50,000$ steps. The network implementation was done using the PyTorch framework. Hyperparameters were set as follows: $\lambda_1=1$, $\lambda_2=1$, and $\lambda_3=1$. In the attention block, three scales of edges extracted by ATW were used. The edge attention module was configured with three basic blocks ($k=3$). Each basic block had four residual blocks ($m=4$), followed by two parallel sets of residual blocks ($n=1$), The ViT block was equipped with sixteen attention heads ($k=16$), a patch size of four ($M=4$), and a multi-layer-perceptron ratio of two. The network maintained a constant channel number of sixty ($C=64$) and utilized a batch size of two during training.

\subsubsection{Comparison} In our comparative analysis, we assessed the performance of our method with $\mathcal{L}_{\ell_1}/\mathcal{L}_{\ell_2}$ loss function alongside several SR techniques, including standard bicubic, DPIR \cite{zhang2021plug}, PSSR \cite{fang2021deep}, and SwinIR \cite{liang2021swinir}, setting hyper-parameters as in the respective papers. As a preprocessing step, we first utilized bicubic interpolation to resize both the LR and HR images to make the closest integer resolution ratio between them. Specifically, we resize the LR and HR images to dimensions of $341 \times 341 \times K$ and $1023 \times 1023 \times K$, leading to resolution ratio  $\tau =3$, where $K$ is number of slices. We conducted comparative experiments using pairs of LR and HR EM images. Additionally, we investigated three training strategies for the proposed method: Training using I) real LR and HR image pairs, II) synthetic LR and HR image pairs, and III) LR and denoised HR image pairs. For synthetic training, LR images are generated using two scenarios. First, by bicubically down-sampling HR images—a common practice in computer vision, we refer to it as synthetic (I). Second, by introducing random isotropic Gaussian kernel ($\kappa \in [0, 3]$) and random zero-mean Gaussian noise ($\sigma \in [20, 40]$) during training, followed by bicubic down-sampling and the addition of random zero-mean Gaussian noise ($\sigma \in [5, 15]$),  we refer to it as synthetic (II).

\subsection{Quality Evaluation Metircs}
To quantitatively assess the effectiveness of the proposed method and compare it with others, we have considered three image quality metrics: the structural similarity index (SSIM) \cite{wang2004image}, peak signal-to-noise ratio (PSNR) as standard metrics, as well as Fourier ring correlation (FRC) \cite{banterle2013fourier}, which is utilized for evaluating EM SR \cite{reid2022multi}.

\subsubsection{SSIM} The SSIM quantifies the similarity between restored $\hat{x}$ and reference $x$ images in terms of luminance, contrast, and structure. It is calculated by:
\begin{equation} \label{eq_18}
\mathrm{SSIM}(x,\hat{x}) = \frac{(2\mu_{x}\mu_{\hat{x}} + c_1)(2\sigma_{x\hat{x}} + c_2)} { (\mu_{x}^2 + \mu_{\hat{x}}^2 + c_1) (\sigma_{x}^2 + \sigma_{\hat{x}}^2 + c_2)},
\end{equation}
where $\mu_{x}$ and $\mu_{\hat{x}}$ are the average pixel intensities of $x$ and $\hat{x}$ (luminance). $\sigma_{x}$ and $\sigma_{\hat{x}}$ are the standard deviations of $x$ and $\hat{x}$ pixel intensities (contrasts), while $\sigma_{x\hat{x}}$ represents the covariance between $x$ and $\hat{x}$ (structural similarity). $c_1$ and $c_2$ are small positive constants for division stability, typically set as $0.01$ and $0.03$ relative to the maximum pixel value, $L$.

\subsubsection{PSNR} {The PSNR measures the ratio of the maximum pixel value to the mean square error (MSE) between the reconstructed image $\hat{x}$ and the ground truth $x$ as below:
\begin{equation}  \label{eq_19}
 \mathrm{PSNR}(x,\hat{x}) =  10\log_{10}  \Bigl(\frac{L^2}{ \mathrm{MSE}(x,\hat{x})}  \Bigl )
\end{equation}    
}

\subsubsection{FRC} {The FRC measures the correlation between reconstructed image $\hat{x}$ and reference $x$ in the frequency domain when spectra $\mathcal{R}$ is subdivided into $N$ concentric rings $r_i$, i.e., $\mathcal{R} = \{r_i\}_{i=1}^N$. FRC is calculated using the following formula:
\begin{equation}
\label{eq_20}
   \mathrm{FRC}(\mathcal{R}) =   \frac{ {\sum_{r_i \in \mathcal{R}} \mathcal F_{x}(r_i)}  \overline {\mathcal F_{\hat{x}}}(r_i) } 
     { \sqrt{ \left({\sum_{r_i \in \mathcal{R}} |\mathcal F_{x}(r_i)|^2 }\right)  \left ( {\sum_{r_i \in \mathcal{R}} |\mathcal F_{\hat{x}}(r_i)|^2} \right )} },
\end{equation}  
where $\mathcal F_{x}(r_i)$ and $\mathcal F_{\hat{x}}(r_i)$ are Fourier transformation of $x$ and $\hat{x}$ over ring $r_i$, and $\mathrm{FRC}(\mathcal{R})$ provides spectral correlation as a function of spatial frequency. The average correlation across the spectra is denoted by $\overline{\mathrm{FRC}}$. 

In the numerical evaluations, the denoised HR 3D-EM images, obtained through the method proposed in \cite{ghahremani2022adversarial}, were utilized as the ground truth references.}

\subsection{Results}
\begin{figure*}[!t]
\centering
\includegraphics[width=0.93\textwidth,height=0.48\textwidth]{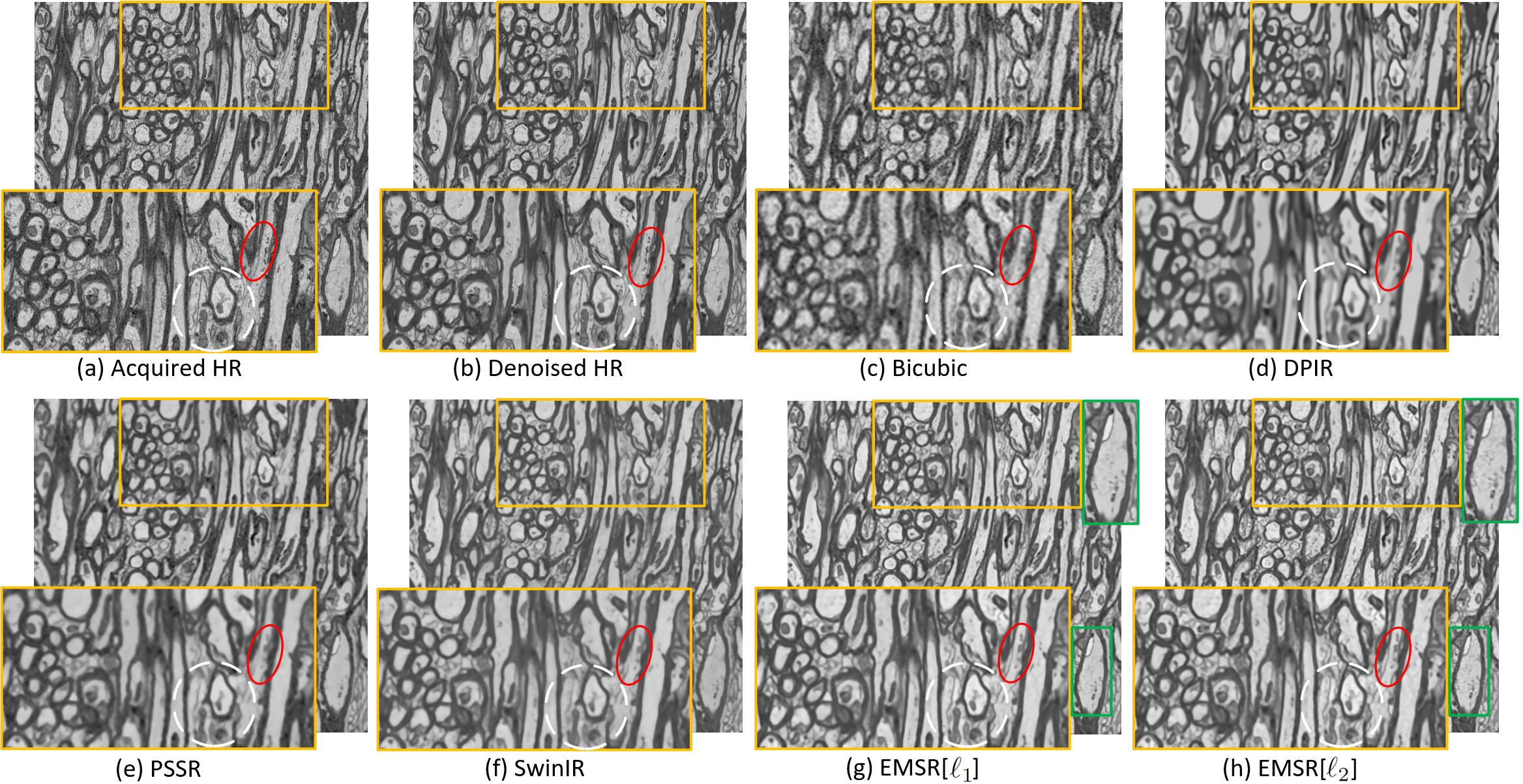}\vspace{-2mm}
\caption{Visual comparisons of super-resolution methods for BRAIN5[IPSI] are presented, and magnified regions are presented to facilitate comparison.}
\label{set5}
\end{figure*}

\begin{figure}[!h]
\centering
\includegraphics[width=0.47\textwidth]{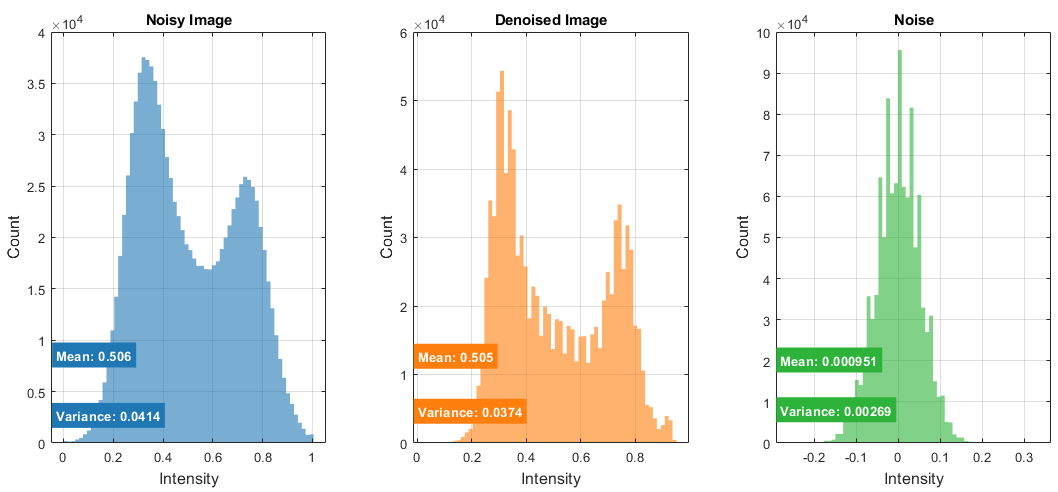}\vspace{-2mm}
\caption{ Conditions for network training using pairs of corrupted images. The histograms of one slice from dataset BRAIN2[contra], its denoised version, and noise are illustrated. The mean and variance presented on each plot are examined to investigate conditions in (\ref{eq_10}). I) $\mathbb{E}[x_{n}] = 5.06  \times 10^{-1}$, $\mathbb{E}[x_{clean}] = 5.05 \times 10^{-1}$, and $\mathbb{E}[n] = 9.51 \times 10^{-4} $, satisfying (\ref{eq_10:I}) that mentions $\mathbb{E}[x_{clean}] \gg \mathbb{E}[n]$. II)  $\sigma_{x_{n}}^2 = 4.14 \times 10^{-2}$, $\sigma_{x_{clean}}^2 = 3.74 \times 10^{-2}$, and $\sigma_{n}^2 = 2.69 \times 10^{-3} $, meeting condition in (\ref{eq_10:II}) that $\sigma_{x_{clean}}^2 \gg \sigma_{n}^2$. It should be noted that here we considered the denoised reference as a clean reference.}
\label{fig_training_condition}
\end{figure}

\begin{figure*}[!t]
\centering
\includegraphics[width=0.93\textwidth,height=0.48\textwidth]{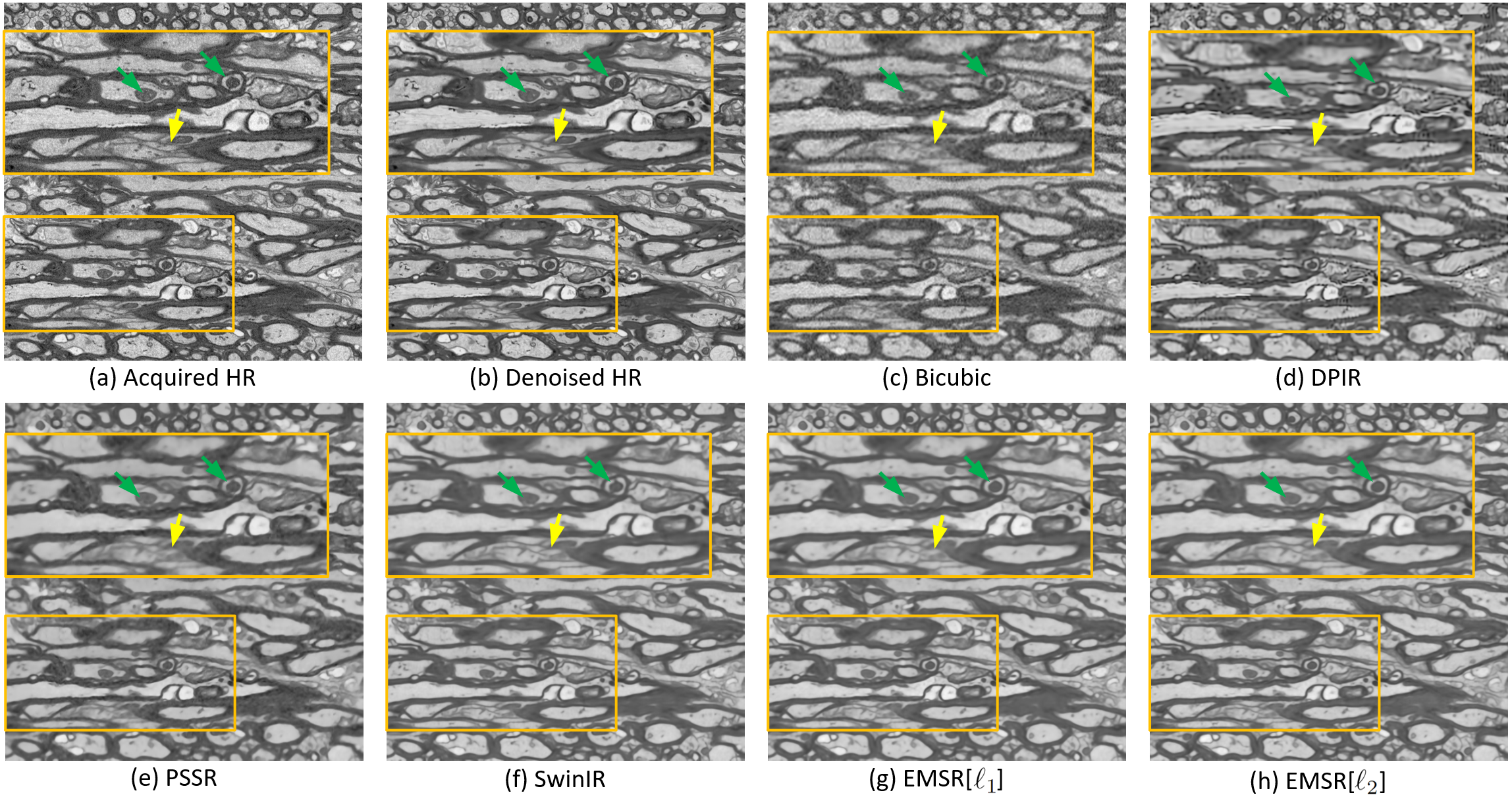}\vspace{-2.5mm}
\caption{Visual comparisons of super-resolution methods for BRAIN2[CONTRA] are presented, and magnified regions are presented to facilitate comparison.}
\label{set2}
\end{figure*}

\begin{figure*}[!t]
\centering
\includegraphics[width=0.93\textwidth ]{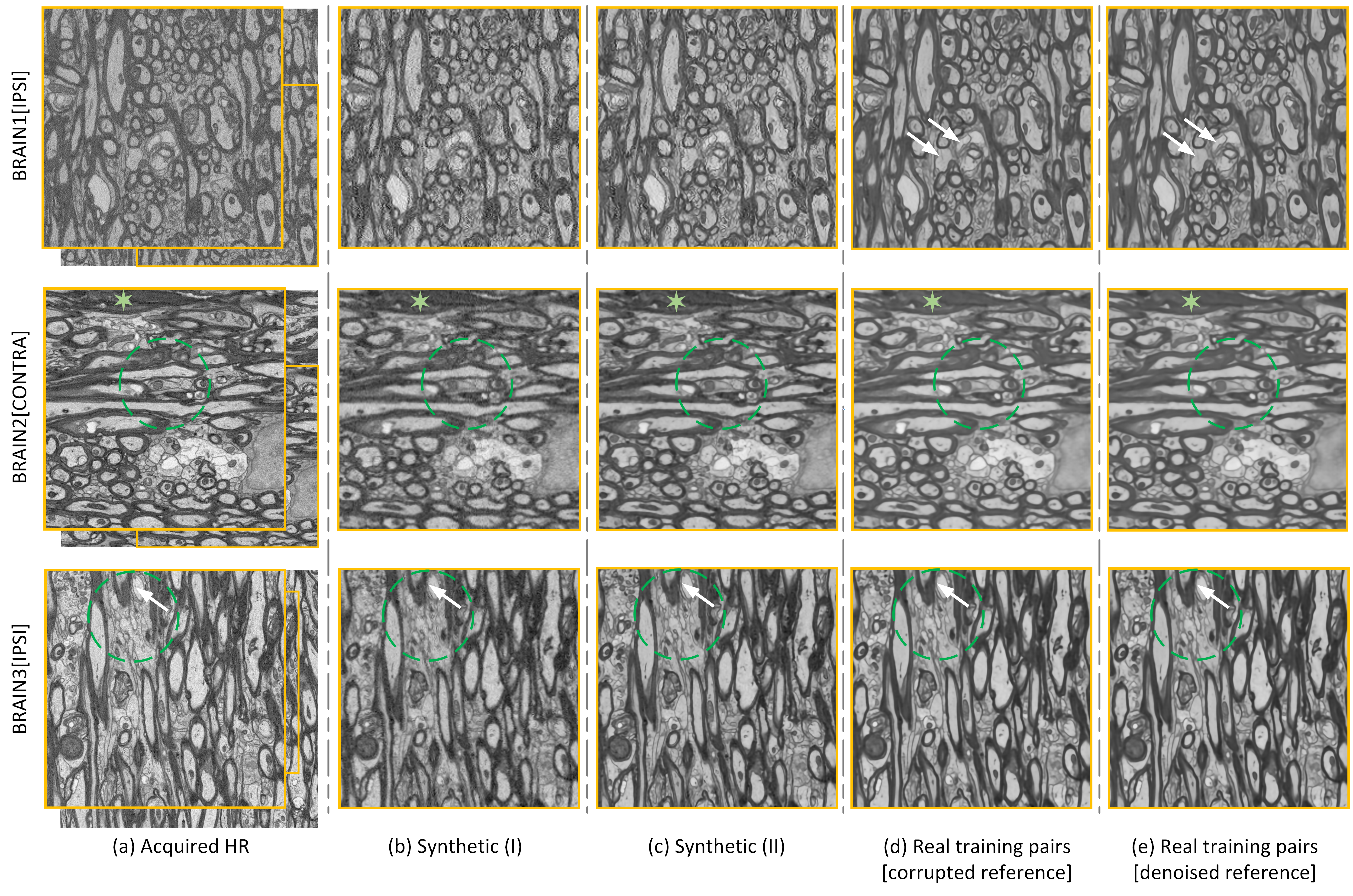}\vspace{-2.5mm}
\caption{Illustrative visual comparisons of EMSR with different data training strategies. Panel (a) displays acquired HR, while (b) and (c) respectively showcase Synthetic (I) and Synthetic (II). Training with real pairs is depicted in both (d) for real noise-like corrupted reference and (e) for denoised reference.}\label{real_noisy_synthetic}
\end{figure*}

\subsubsection{Method Comparison}
The comparative results were obtained through a five-fold cross-validation process, where data from one animal functioned as test sets, and data from other animals were used as training sets. The quantitative results are summarized in Table (\ref{tab1}). The reported average values, based on SSIM, emphasize the inferior performance of the bicubic method compared to deep learning-based methods—DPIR, PSSR, SwinIR, and EMSR. Among these methods, SwinIR and EMSR[ours] obtained superior quality metrics. Our approach, employing the ${\ell_1}$ and ${\ell_2}$ loss function, showcased the highest and second-highest scores, respectively. Similarly, the reported FRC value, demonstrates the superior performance of EMSR in terms of spectral correlation between restored and ground truth images compared to the competitors, achieving the best and second-best scores,  when respectively trained with ${\ell_1}$ and ${\ell_2}$ loss functions. However, in terms of PSNR, DPIR achieved the highest score, and the PSSR method achieved the second-best PSNR. It's essential to emphasize that the effectiveness of PSNR as an evaluation metric for SR model performance is limited. This limitation arises from its pure reliance on pixel values and its inability to capture a direct structural correlation between super-resolved and ground truth images.

To conserve space, we present a curated selection of representative results in figures (\ref{set5}) and (\ref{set2}). These figures provide visual insights into scenarios where our proposed method excelled as the best and also where it did not attain the highest quantitative performance.

Figure \ref{set5} showcases results from the BRAIN5[IPSI]. In this sub-dataset, our proposed method demonstrated outstanding performance, achieving the best and second-best quantitative results, based on SSIM and FRC, when utilizing $\ell_1$ and $\ell_2$ loss functions, respectively. In panel (a), we observe the bicubically interpolated LR image, which exhibits a lack of visual clarity and maintains noise. Conversely, DL-based SR methods effectively reduce noise, as evident in Fig. \ref{set5} (d)-(h). Among these methods, DPIR, i.e., a PnP method, produces overly smooth results, particularly when restoring fine details, as shown within regions enclosed by the ellipsoid and dashed circle. This outcome can be attributed to mismatches between priors in the trained model and test images. In contrast, PSSR, SwinIR, and EMSR, which were trained using EM images, exhibit the capability to restore intricate details and nuances characteristic of EM brain images. Among them, PSSR sometimes failed to restore particular intricate edges, as represented by the area confined by an ellipsoid. It also led to smear-out edges, as indicated within the dashed circle. Similarly, SwinIR faced challenges in recovering certain edges, akin to PSSR, showcasing within the region confined by the ellipsoid. It also introduced blurred output and fuzzy edges within an area marked by the dashed circle. On the other hand, EMSR with both $\ell_1$ and $\ell_2$ loss functions successfully super-resolved LR images by restoring intricate edges with higher contrast while avoiding blurriness. When comparing the results between $\ell_1$ and $\ell_2$, $\ell_1$ exhibited slightly better noise suppression (see zoomed-in rectangle marked in green). These results align with the theory mentioning that $\ell_1$ loss, in contrast to $\ell_2$, does not over-penalize large errors, resulting in fewer noise artifacts. The condition checking for training with a no-clean reference is depicted in Figure \ref{fig_training_condition}.

Figure \ref{set2} presents results from BRAIN2[CONTRA]. In this subset, the SwinIR method exhibited superior performance in SSIM and FRC, collectively indicating enhanced structural capabilities. SwinIR did not achieve the highest PSNR, yet it maintained a satisfactory level of pixel fidelity. Panel (c) and (d) show that bicubic and DPIR generally produced oversmooth details, as denoted by the yellow arrow. Panel (e) revealed that PSSR excelled in enhancing details and contrast but faced challenges in recovering fine edges, as indicated by the yellow arrow. SwinIR and EMSR (f)-(h) showcased superior resolution enhancement and noise reduction. Particularly, SwinIR delivered slightly sharper SR results, highlighted by the yellow arrow. However, the proposed method, likely due to its edge-attention mechanism, demonstrated a superior ability to super-resolve two closely situated compartments compared to SwinIR, which struggled to effectively separate them, pointed by the green arrows.

\subsubsection{Data Training Strategies}
The outcomes of training with different strategies—real pairs featuring corrupted references, real pairs with a denoised reference, and synthetic LR and HR pairs (both synthetic (I) and (II))—are detailed in Table (\ref{tab2_temp}). The reported average quantitative results across all datasets revealed that training with an acquired HR image as a reference and its denoised version resulted in nearly identical SSIM and FRC values, with the denoised reference exhibiting an inferior PSNR. Additionally, it was noted that training with synthetic (I) did not attain favorable super-resolution results. In contrast, synthetic (II) exhibited varying performance with promising outcomes, the average performance was slightly lower than that achieved with real pairs.

Representative results are depicted in Figure \ref{real_noisy_synthetic}, spanning from inferior to superior performance. The results for BRAIN1[IPSI] indicate that the trained network for both synthetic (I) and (II), failed to produce satisfactory super-resolution results, as evident from different artifacts. The underlying reason for these shortcomings lies in the inability of both bicubically down-sampling and a pool of random Gaussian noise and blurring kernels to effectively match the degradation in the input LR image. Furthermore, it demonstrates that training using real pairs with either corrupted or denoised references yielded nearly identical outputs, with only subtle differences, such as slightly more homogeneous areas in the case of training with denoised reference, see white arrows. 

\begin{figure}[!t]
\centering
\includegraphics[width=0.45\textwidth,height=0.31\textwidth]{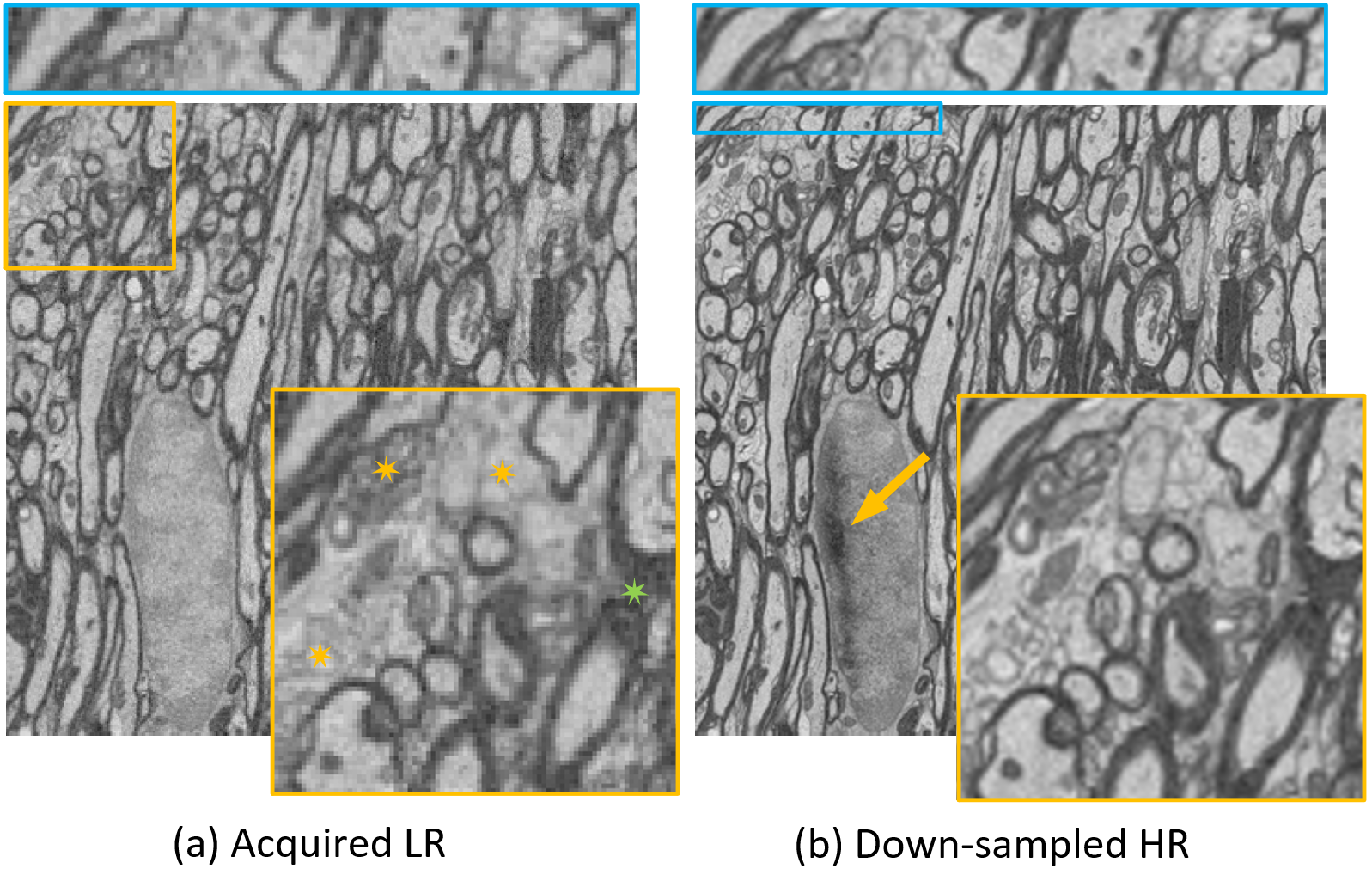}\vspace{-3mm}
\caption{A comparison is shown between a physically acquired LR EM image and a synthetically generated LR image obtained by down-sampling an HR EM image. Notable differences are observed in terms of fine details and intensity. The blue rectangle highlights a drift distortion at the border of the HR image, which is not present in the LR counterpart. An arrow points to a charging effect that is only seen in the HR image. The zoomed-in area accentuates the distinct difference in fine details (yellow asterisks), and intensity level (green asterisk).}\label{Fig_synthetic_VS_acquiredLR}
\end{figure}

The results for BRAIN2[CONTRA] indicate encouraging findings. The Synthetic (I) training strategy yields unsatisfactory results as it struggles to match the degradations present in the input LR image, see Figure \ref{Fig_synthetic_VS_acquiredLR}. However, results for synthetic (II), training with a diverse range of degradations, outperform training with real pairs, generating sharper edges and enhanced contrast, as indicated by the dashed green circle. The key factor behind these results is synthetic training's ability, under well-matched degradations, to learn deblurring and denoising while super-resolving the input LR image. The low-level feature fidelity in the synthetic pairs is well-preserved compared to training with acquired LR and HR images, even in the case of synthetic (I) with bicubic downsampling, evident in black areas marked with asterisks. From a denoising perspective, training with real pairs may offer better performance, benefiting from the independence of noise-like corruptions in independently acquired LR and HR images, preventing the learning of noise-like patterns with random characteristics. Notably, both noisy and denoised reference training produce similar outputs.

BRAIN3[IPSI] showcases additional promising outcomes with the synthetic (II) strategy, demonstrating superior super-resolution performance in recovering fine details, and achieving sharp edges while mitigating noise—highlighted in areas marked by circles and arrows.

When comparing real and synthetic datasets, it is recommended to use real image pairs as they have the potential to enhance the overall quality. The foremost advantage is learning real degradations, which are difficult to simulate, see Fig.\ref{Fig_synthetic_VS_acquiredLR}. Importantly, the separate acquisition of LR and HR images leads to nearly independent noise-like corruption. This independence is beneficial for the network as it prevents the learning of noise-like patterns with random characteristics, learning to denoise while super-resolving LR image. Furthermore, the results indicate that while pairs of acquired synthetic LR, derived from down-sampled HR, and HR images are not suitable as training pairs, there is a potential for computationally generated pairs to advance EM super-resolution. Notably, this approach can address mismatches between acquired LR and HR pairs, i.e., co-registration and contrast, reduce imaging time, and lower costs.

\subsubsection{Super-Resolver as Enhancer}
\begin{figure}[!t]
\centering
\includegraphics[width=0.45\textwidth]{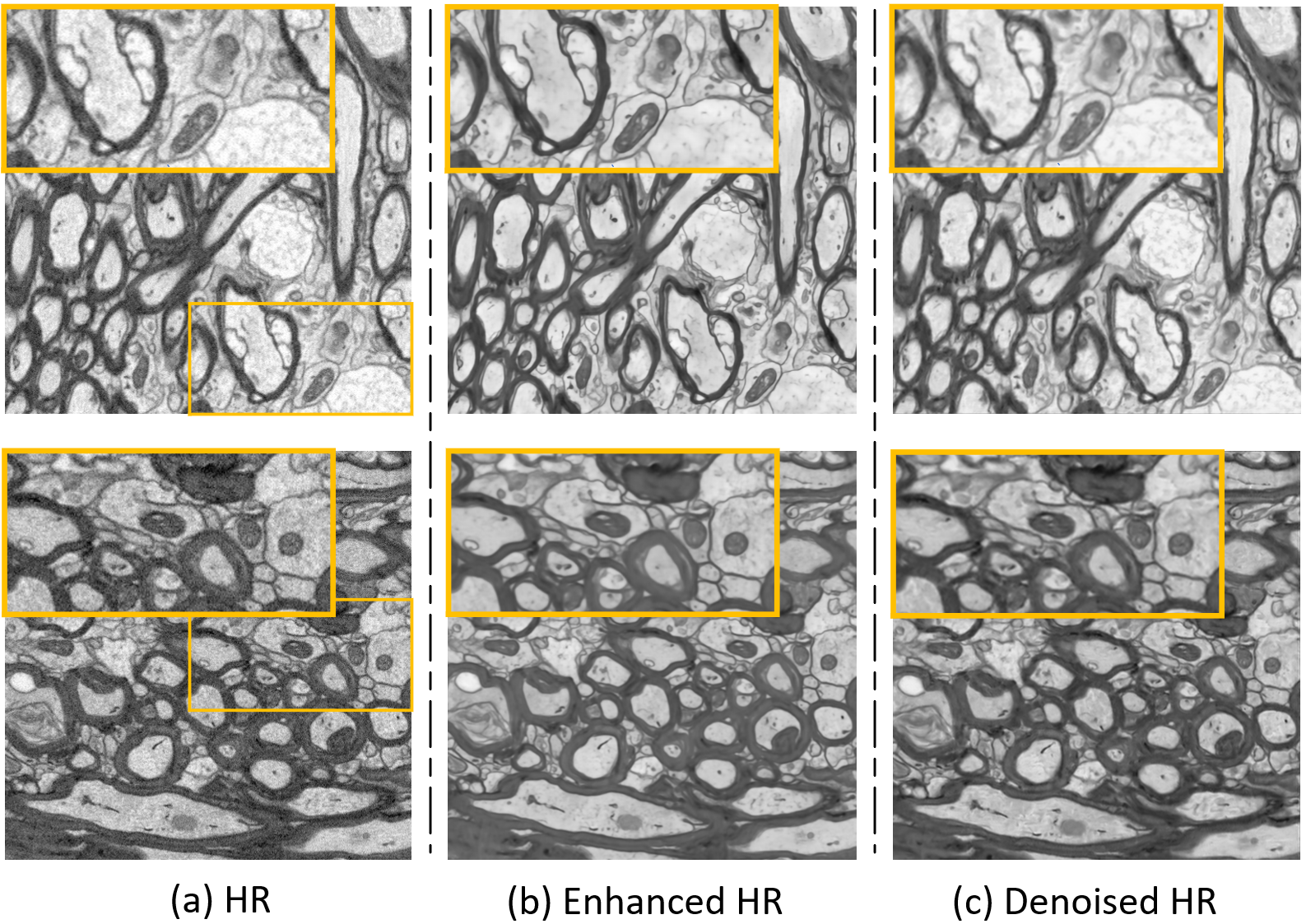}\vspace{-3mm}
\caption{Application of the trained super-resolution model to HR images with the same resolution as the HR images used for training. (a) Input HR, (b) Enhanced using super-resolver, and (c) Denoised HR.}\label{sr_as_denoiser}
\end{figure}

Applying the trained SR model to HR images with the same resolution enhances image quality. In comparison to a denoiser, it enhances the resolution as well as mitigating noise, see first row in Figure \ref{sr_as_denoiser}. However, in situations where there are mismatches between the trained model and the input image, changes in image contrasts may occur, as depicted in the second row of Figure \ref{sr_as_denoiser}. This observation highlights the potential of SR methods to function as denoisers and enhancers, particularly emphasizing the practical capabilities of a self-supervised SR approach that can address the mismatches.

\subsubsection{Super-Resolution can help Distortion Avoidance}
\begin{figure}[!t]
\centering
\includegraphics[width=0.45\textwidth,height=0.35\textwidth]{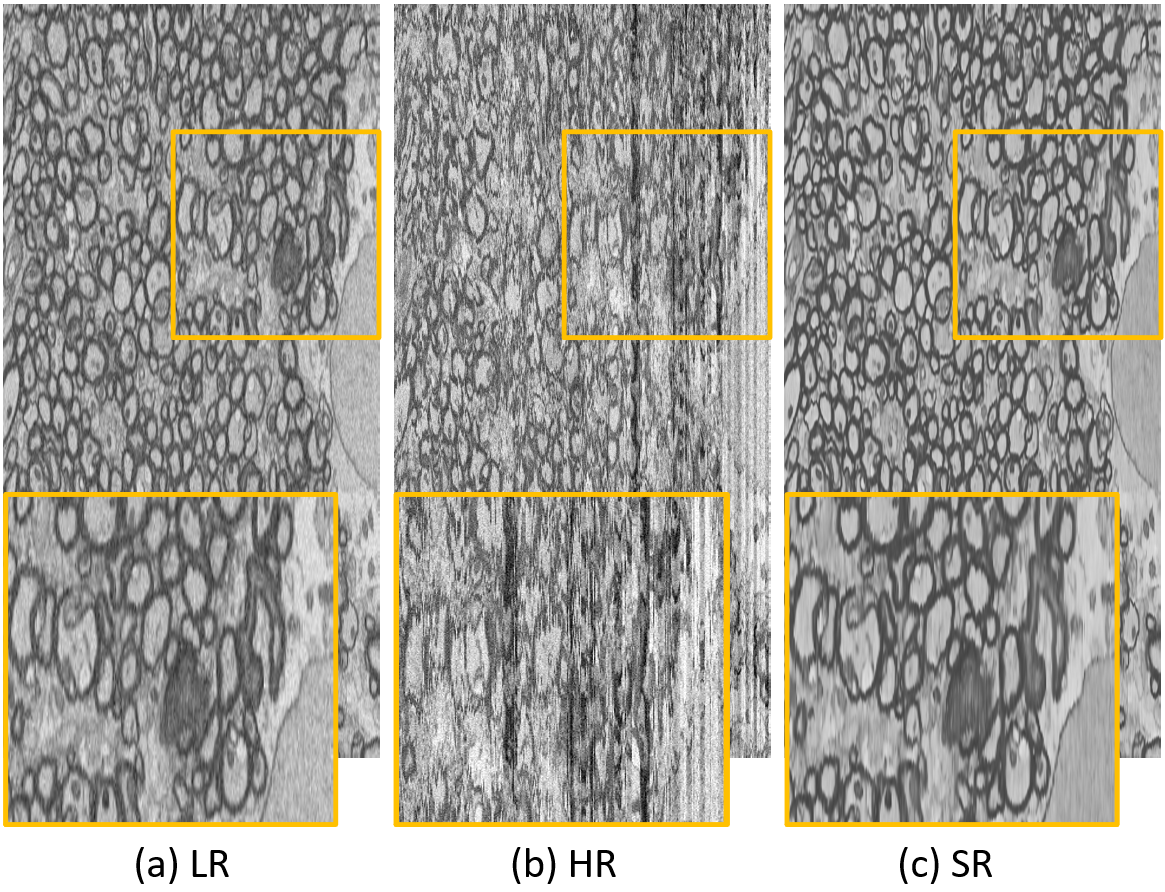}\vspace{-3mm}
\caption{Distortion in the border of 3D-EM data. The $xz$-perspective view of (a) bicubically interpolated acquired LR,  (b) acquired HR, and (c) super-resolved LR images. }\label{Fig_distortion}
\end{figure}

EM imaging at HR may result in distortions at the image border in the $xy$-plane, a phenomenon not observed in LR imaging, as depicted in Figure \ref{Fig_distortion}. However, employing SR techniques enables the generation of an HR image from an LR image, effectively overcoming these distortions.

\subsubsection{Natural Image Pre-trained Networks on Brain EM}
\begin{figure}[!t]
\centering
\includegraphics[width=0.45\textwidth,height=0.35\textwidth]{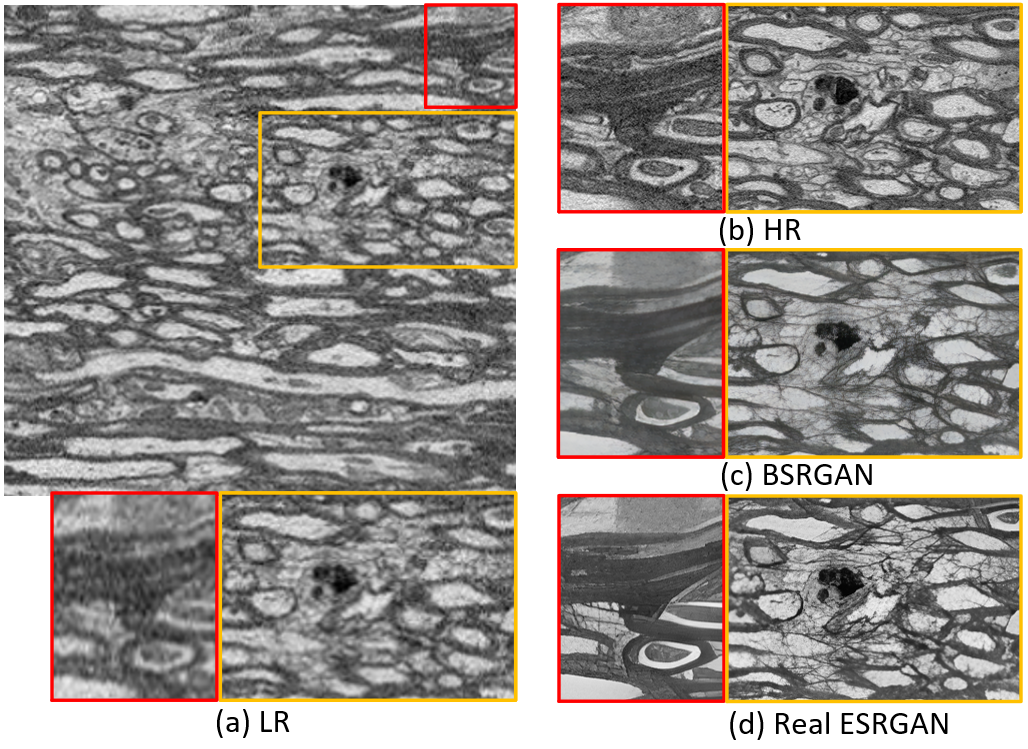}\vspace{-3mm}
\caption{ Super-resolution of EM images using state-off-the-art pre-trained networks designed for natural images. (a) LR, (b) HR, (c) BSRGAN, and (d) Real ESRGAN. }\vspace{-2mm}\label{Traning_natural_images}
\end{figure}

Figure \ref{Traning_natural_images} depicts the application of state-of-the-art pre-trained networks designed for natural images on brain EM. BSRGAN \cite{zhang2021designing} and Real ESRGAN \cite{wang2021real} are two networks designed for the super-resolution of natural images–which respectively were trained on natural and pure synthetic datasets. When applied to brain EM images, while these methods can restore the overall structure of large tissue compartments, they fail to recover the intricate details and nuances unique to brain EM. In particular, they tend to introduce unrealistic
details and cartoonish textures, as visible in the zoomed-in areas.

\begin{table*}[t]
\begin{center}
\centering \vspace{-5mm}
\caption{Quantitative evaluation of super-resolution methods. The best and second-best scores are in bold and underlined, respectively. Reported values for mean and standard deviation ($\mu\pm\sigma$) for each BRAIN are calculated across all slices. The overall evaluation, highlighted in gray, represents the mean and standard deviation across reported mean values for all datasets.}\label{tab1}\vspace{-1.5mm}
\setlength{\tabcolsep}{2.1pt} 
\begin{tabular}{|c|l||c|c|c|c|c|c|}
\hline
\textbf{Metric} & \multicolumn{1}{c||}{\textbf{Datasets}} & \textbf{Bicubic} & \textbf{DPIR} & \textbf{PSSR} & \textbf{SwinIR} & \textbf{EMSR [OURS /$\ell_1$]} & \textbf{EMSR [OURS /$\ell_2$]} \\
\hline \hline   
\multirow{10}{*}{\rotatebox[origin=c]{90}{\textbf{SSIM}} $\Large\uparrow$}                                                                                            
& \text{BRAIN1 [IPSI]}    &$0.551\pm0.012$ &$\itunderT{0.712}{0.015}$ &$0.702\pm0.014$  &$0.697\pm0.015$ &$\boldT{0.720}{0.015}$ &$0.705\pm0.015$\\
& \text{BRAIN1 [CONTRA]}  &$0.719\pm0.013$ &$\itunderT{0.840}{0.014}$ &$\boldT{0.849}{0.013}$  &$0.832\pm0.015$ &$0.832\pm0.015$ &$0.815\pm0.016$\\
& \text{BRAIN2 [IPSI]}    &$0.669\pm0.014$ &$0.715\pm0.014$ &$0.713\pm0.014$  &$0.694\pm0.014$ &$\boldT{0.739}{0.016}$ &$\itunderT{0.737}{0.014}$\\
& \text{BRAIN2 [CONTRA]}  &$0.640\pm0.020$ &$0.669\pm0.024$ &$0.672\pm0.026$  &$\boldT{0.695}{0.025}$ &$\itunderT{0.688}{0.026}$ &$\boldT{0.695}{0.025}$\\
& \text{BRAIN3 [IPSI]}    &$0.646\pm0.008$ &$0.673\pm0.008$ &$0.666\pm0.010$  &$\itunderT{0.715}{0.009}$ &$\itunderT{0.715}{0.009}$ &$\boldT{0.720}{0.009}$\\
& \text{BRAIN3 [CONTRA]}  &$0.737\pm0.026$ &$0.753\pm0.030$ &$0.740\pm0.031$  &$\itunderT{0.787}{0.026}$ &$0.780\pm0.028$ &$\boldT{0.792}{0.026}$\\
& \text{BRAIN4 [IPSI]}    &$0.721\pm0.007$ &$0.794\pm0.014$ &$\itunderT{0.808}{0.012}$  &$0.790\pm0.007$ &$\boldT{0.809}{0.009}$ &$0.796\pm0.008$\\
& \text{BRAIN4 [CONTRA]}  &$0.684\pm0.014$ &$0.717\pm0.016$ &$0.728\pm0.018$  &$\boldT{0.745}{0.018}$ &$0.735\pm0.020$ &$\itunderT{0.738}{0.018}$\\
& \text{BRAIN5 [IPSI]}    &$0.615\pm0.019$ &$0.639\pm0.016$ &$0.662\pm0.016$  &$0.669\pm0.018$ &$\boldT{0.687}{0.018}$ &$\itunderT{0.681}{0.017}$\\
& \cellcolor{gray!30} \textbf{ALL DATASETS} &\cellcolor{gray!30}$0.665\pm0.059$ &\cellcolor{gray!30}$0.724\pm0.064$ &\cellcolor{gray!30}$0.727\pm0.065$ &\cellcolor{gray!30}$0.736\pm0.056$ &\cellcolor{gray!30}$\boldT{0.745}{0.051}$ &\cellcolor{gray!30}$\itunderT{0.742}{0.048}$ \\ \hline \hline
\multirow{10}{*}{\rotatebox[origin=c]{90}{\textbf{PSNR}} $\Large\uparrow$}
& \text{BRAIN1 [IPSI]}    &$23.1\pm0.3$ &$24.8\pm0.4$ &$\boldT{25.0}{0.4}$  &$24.4\pm0.4$ &$\itunderT{24.9}{0.4}$ &$\itunderT{24.9}{0.4}$\\
& \text{BRAIN1 [CONTRA]}  &$24.8\pm2.6$ &$\itunderT{25.9}{2.9}$ &$\boldT{26.1}{3.2}$  &$24.4\pm2.9$ &$24.2\pm2.7$ &$23.8\pm2.6$\\
& \text{BRAIN2 [IPSI]}    &$23.1\pm1.5$ &$\boldT{23.6}{1.7}$ &$\itunderT{23.6}{2.0}$  &$22.2\pm1.2$ &$22.9\pm1.6$ &$23.0\pm1.6$\\
& \text{BRAIN2 [CONTRA]}  &$22.8\pm0.9$ &$\itunderT{23.3}{1.0}$ &$\boldT{23.3}{0.9}$  &$22.5\pm1.6$ &$22.3\pm1.5$ &$22.3\pm1.5$\\
& \text{BRAIN3 [IPSI]}    &$21.9\pm0.7$ &$22.2\pm0.8$ &$21.8\pm0.8$  &$\boldT{23.0}{0.8}$ &$22.7\pm0.7$ &$\itunderT{22.8}{0.7}$\\
& \text{BRAIN3 [CONTRA]}  &$21.8\pm1.3$ &$22.1\pm1.3$ &$21.5\pm1.3$  &$\itunderT{23.0}{1.4}$ &$22.7\pm1.5$ &$\boldT{23.1}{1.4}$\\
& \text{BRAIN4 [IPSI]}    &$24.3\pm0.4$ &$\itunderT{25.2}{0.5}$ &$\itunderT{25.2}{0.5}$  &$\boldT{25.3}{0.4}$ &$24.2\pm0.6$ &$23.3\pm0.5$\\
& \text{BRAIN4 [CONTRA]}  &$24.0\pm0.7$ &$24.4\pm0.8$ &$\boldT{24.6}{0.8}$  &$22.1\pm0.9$ &$24.0\pm0.7$ &$\itunderT{24.4}{0.6}$\\
& \text{BRAIN5 [IPSI]}    &$22.6\pm0.6$ &$\itunderT{22.7}{0.7}$ &$\boldT{23.2}{0.7}$  &$22.5\pm0.7$ &$21.6\pm1.2$ &$21.5\pm1.2$\\
& \cellcolor{gray!30} \textbf{ALL DATASETS} &\cellcolor{gray!30}$23.2\pm1.0$ &\cellcolor{gray!30}$\boldT{23.8}{1.4}$ &\cellcolor{gray!30}$\itunderT{23.8}{1.6}$ &\cellcolor{gray!30}$23.3\pm1.1$ &\cellcolor{gray!30}$23.3\pm1.1$ &\cellcolor{gray!30}$23.2\pm1.0$ \\ \hline \hline
\multirow{10}{*}{\rotatebox[origin=c]{90}{\boldmath{$\overline{\mathrm{FRC}}$}} $\Large\uparrow$}
& \text{BRAIN1 [IPSI]}    &$0.191\pm0.004$ &$0.216\pm0.008$ &$0.227\pm0.007$  &$0.243\pm0.010$ &$\boldT{0.253}{0.011}$ &$\itunderT{0.246}{0.010}$\\
& \text{BRAIN1 [CONTRA]}  &$0.198\pm0.008$ &$0.210\pm0.008$ &$0.244\pm0.007$  &$\itunderT{0.281}{0.014}$ &$\boldT{0.285}{0.013}$ &${0.280}\pm{0.013}$\\
& \text{BRAIN2 [IPSI]}    &$0.226\pm0.007$ &$0.287\pm0.012$ &$0.287\pm0.012$  &$0.310\pm0.013$ &$\boldT{0.319}{0.015}$ &$\itunderT{0.317}{0.013}$\\
& \text{BRAIN2 [CONTRA]}  &$0.213\pm0.009$ &$0.220\pm0.010$ &$0.282\pm0.013$  &$\boldT{0.314}{0.016}$ &$\itunderT{0.307}{0.014}$ &$0.304\pm0.014$\\
& \text{BRAIN3 [IPSI]}    &$0.245\pm0.004$ &$0.246\pm0.006$ &$0.309\pm0.006$  &$0.345\pm0.008$ &$\boldT{0.352}{0.008}$ &$\itunderT{0.349}{0.008}$\\
& \text{BRAIN3 [CONTRA]}  &$0.259\pm0.007$ &$0.263\pm0.009$ &$0.319\pm0.010$  &$\itunderT{0.350}{0.016}$ &$0.349\pm0.015$ &$\boldT{0.352}{0.015}$\\
& \text{BRAIN4 [IPSI]}    &$0.216\pm0.004$ &$0.220\pm0.008$ &$0.273\pm0.009$  &$0.292\pm0.010$ &$\boldT{0.297}{0.009}$ &$\itunderT{0.297}{0.010}$\\
& \text{BRAIN4 [CONTRA]}  &$0.261\pm0.005$ &$0.256\pm0.008$ &$\itunderT{0.342}{0.010}$  &$\boldT{0.349}{0.010}$ &$0.340\pm0.012$ &$0.337\pm0.011$\\
& \text{BRAIN5 [IPSI]}    &$0.227\pm0.007$ &$0.222\pm0.008$ &$0.290\pm0.010$  &$0.311\pm0.010$ &$\boldT{0.323}{0.011}$ &$\itunderT{0.320}{0.011}$\\
& \cellcolor{gray!30} \textbf{ALL DATASETS} &\cellcolor{gray!30}$0.226\pm0.025$ &\cellcolor{gray!30}$0.238\pm0.026$ &\cellcolor{gray!30}$0.286\pm0.036$ &\cellcolor{gray!30}$0.311\pm0.035$ &\cellcolor{gray!30}$\boldT{0.314}{0.032}$ &\cellcolor{gray!30}$\itunderT{0.311}{0.034}$ \\ \hline 
\end{tabular}
\end{center}
\end{table*}

\begin{table*}
\centering \vspace{-5mm}
\caption{Quantitative evaluation of EMSR using different training strategies: pairs of real LR and HR, pairs of synthetic LR and HR, and pairs of real LR and denoised HR, for $\ell_1$ loss function. The best and second-best scores are in bold and underlined, respectively. Reported values for mean and standard deviation ($\mu\pm\sigma$) for each BRAIN are calculated across all slices.}\label{tab2_temp}\vspace{-1.5mm}
\resizebox{2\columnwidth}{!}{%
\begin{tabular}{| c | l || c | c | c | c | c | c | c | c | c |c|}
\hline
\textbf{Metric} & \multicolumn{1}{c||}{\textbf{Method}} & \multicolumn{2}{c|}{\textbf{BRAIN1}} & \multicolumn{2}{c|}{\textbf{BRAIN2}} & \multicolumn{2}{c|}{\textbf{BRAIN3}} & \multicolumn{2}{c|}{\textbf{BRAIN4}} & \textbf{BRAIN5}   &\cellcolor{gray!30}\textbf{ALL DATASETS}\\
\cline{3-11}
& & \textbf{IPSI} & \textbf{CONTRA} & \textbf{IPSI} & \textbf{CONTRA} & \textbf{IPSI} & \textbf{CONTRA} & \textbf{IPSI} & \textbf{CONTRA} & \textbf{IPSI} &\cellcolor{gray!30} \\
\hline \hline 
\multirow{3}{*}{\rotatebox[origin=c]{90}{\textbf{SSIM}} $\Large\uparrow$}  
& \text{EMSR[Real]}           &$\itunderT{0.720}{0.015}$  &$\itunderT{0.832}{0.015}$  &$\itunderT{0.739}{0.016}$ &$\itunderT{0.688}{0.026}$  &$\itunderT{0.715}{0.009}$  &$\itunderT{0.780}{0.028}$   &$\boldT{0.809}{0.009}$      &$\itunderT{0.735}{0.020}$   &$\boldT{0.687}{0.018}$      &\cellcolor{gray!30}$\boldT{0.745}{0.051}$ \\ 
& \text{EMSR[Denoised]}       &$\boldT{0.729}{0.016}$  &$\boldT{0.839}{0.015}$  &$\boldT{0.740}{0.017}$ &$0.685\pm0.026$  &$0.712\pm0.009$  &$0.772\pm0.028$   &$\itunderT{0.808}{0.009}$      &$0.734\pm0.020$   &$\itunderT{0.686}{0.018}$      &\cellcolor{gray!30}$\itunderT{0.745}{0.053}$ \\
& \text{EMSR[Synthetic (II)]} &$0.598\pm0.013$  &$0.780\pm0.014$  &$0.738\pm0.014$ &$\boldT{0.704}{0.021}$  &$\boldT{0.724}{0.009}$  &$\boldT{0.794}{0.025}$   &$0.776\pm0.006$      &$\boldT{0.755}{0.018}$   &$0.667\pm0.021$      &\cellcolor{gray!30}$0.726\pm0.063$ \\
& \text{EMSR[Synthetic (I)]}  &$0.519\pm0.012$  &$0.705\pm0.015$  &$0.675\pm0.013$ &$0.645\pm0.020$  &$0.663\pm0.008$  &$0.749\pm0.025$   &$0.718\pm0.008$      &$0.708\pm0.015$   &$0.625\pm0.020$      &\cellcolor{gray!30}$0.667\pm0.068$ \\ \hline
\multirow{3}{*}{\rotatebox[origin=c]{90}{\textbf{PSNR}} $\Large\uparrow$}  
& \text{EMSR[Real]}           &$\itunderT{24.9}{0.4}$  &$24.2\pm2.7$  &$22.9\pm1.6$ &$22.3\pm1.5$  &$\itunderT{22.7}{0.7}$  &$\boldT{22.7}{1.5}$   &${24.2}\pm{0.6}$      &$24.0 \pm 0.7$   &$21.6\pm1.2$      &\cellcolor{gray!30}$\itunderT{23.3}{1.1}$ \\  
& \text{EMSR[Denoised]}       &$\boldT{25.0}{0.4}$  &${24.5}\pm{2.8}$  &$\itunderT{23.4}{1.6}$ &$22.6\pm1.5$  &$\itunderT{22.7}{0.7}$  &$22.1\pm1.5$   &$22.9\pm0.7$      &$23.6\pm0.7$   &$21.5\pm1.2$      &\cellcolor{gray!30}$23.1\pm1.1$ \\
& \text{EMSR[Synthetic (II)]} &$23.3\pm0.3$  &$\boldT{25.0}{2.6}$  &$\boldT{24.0}{1.9}$ &$\boldT{23.8}{1.0}$  &$\boldT{22.9}{0.9}$  &$\itunderT{22.4}{1.4}$   &$\boldT{24.7}{0.6}$      &$\boldT{24.9}{0.9}$   &$\boldT{22.8}{0.7}$      &\cellcolor{gray!30}$\boldT{23.8}{1.0}$ \\
& \text{EMSR[Synthetic (I)]}  &$22.6\pm0.3$  &$\itunderT{24.7}{2.5}$  &$23.1\pm1.6$ &$\itunderT{22.9}{0.9}$  &$22.0\pm0.8$  &$21.8\pm1.3$   &$\itunderT{24.2}{0.4}$      &$\itunderT{24.3}{0.7}$   &$\itunderT{22.7}{0.6}$      &\cellcolor{gray!30}$23.1\pm1.0$ \\  \hline \hline
\multirow{3}{*}{\rotatebox[origin=c]{90}{\boldmath{$\overline{\mathrm{FRC}}$}} $\Large\uparrow$}  
& \text{EMSR[Real]}           &$\itunderT{0.253}{0.011}$  &$\itunderT{0.285}{0.013}$  &$\itunderT{0.319}{0.015}$ &$\itunderT{0.307}{0.014}$  &$\itunderT{0.352}{0.008}$  &$\itunderT{0.349}{0.015}$   &$\boldT{0.297}{0.010}$      &$\itunderT{0.340}{0.012}$   &$\boldT{0.323}{0.011}$      &\cellcolor{gray!30}$\itunderT{0.314}{0.032}$ \\ 
& \text{EMSR[Denoised]}       &$\boldT{0.256}{0.011}$  &$\boldT{0.288}{0.013}$  &$\boldT{0.321}{0.015}$ &$\itunderT{0.307}{0.014}$  &${0.349}\pm{0.009}$  &$0.347\pm0.016$   &$\itunderT{0.294}{0.010}$      &$\itunderT{0.340}{0.012}$   &$\itunderT{0.320}{0.012}$      &\cellcolor{gray!30}$\boldT{0.314}{0.031}$ \\ 
& \text{EMSR[Synthetic (II)]} &$0.214\pm0.008$  &$0.255\pm0.015$  &$0.311\pm0.015$ &$\boldT{0.311}{0.015}$  &$\boldT{0.367}{0.010}$  &$\boldT{0.362}{0.018}$   &$0.289\pm0.012$      &$\boldT{0.369}{0.013}$   &$0.308\pm0.015$      &\cellcolor{gray!30}$0.310\pm0.053$ \\
& \text{EMSR[Synthetic (I)]}  &$0.193\pm0.005$  &$0.205\pm0.009$  &$0.245\pm0.010$ &$0.235\pm0.013$  &$0.285\pm0.006$  &$0.294\pm0.011$   &$0.230\pm0.006$      &$0.313\pm0.009$   &$0.249\pm0.010$      &\cellcolor{gray!30}$0.250\pm0.040$ \\  \hline
\end{tabular}%
}
\end{table*}

\section{Conclusion}
We introduced a deep-learning-based SR framework named EMSR, to address the challenge of acquiring clean HR 3D-EM images across large tissue volumes. As corruptions are inherent in EM, training neural networks with no-clean references for $\ell_2$ and $\ell_1$ loss functions was explored. Following this, we crafted a noise-robust network that integrated both edge-attention and self-attention mechanisms, to focus on enhancing edge features over less informative backgrounds in brain EM images. Utilizing real LR and HR brain EM image pairs, the network underwent training with LR and HR pairs, along with LR and denoised HR pairs. Experimental results, in line with the discussed theory, confirmed the feasibility of training with no-clean references for both loss functions. While both losses demonstrated similar SR performance, consistent with the literature, $\ell_1$ slightly outperformed $\ell_2$. Furthermore, EMSR demonstrated superior or competitive results, both quantitatively and qualitatively, when compared to established SR methods. In addition to training with real LR and HR pairs, we synthesized LR images from HR using a wide-ranging isotropic Gaussian noise and Gaussian kernels. Experiments with synthetic pairs showed promising results, comparable to models trained on real pairs. Notably, in some cases, the synthesis produced super-resolved images with sharper edges and improved contrasts, addressing inherent mismatches in LR and HR pairs, e.g., co-registration and contrast. This synthesis could also aid in deblurring while denoising and super-resolving LR EM.

EMSR offers improved resolution and reduced noise simultaneously, enabling the computational generation of clean HR EM images over large samples from cost-effective LR EM imaging, providing it as a neuroimaging preprocessing tool for visualization and analysis. 

\section*{Acknowledgments}
The authors would like to thank the Electron Microscopy Unit at the Institute of Biotechnology, University of Helsinki, Finland, for 3D-EM datasets. They would also thank the Bioinformatics Center at the University of Eastern Finland, Finland, and the CSC–IT Center for Science, Finland, for providing computational resources.

\appendices
\section{Training using corrupted reference}
Let $\hat{x}$ and $x$ be random variables such that $\hat{x} = x + n$, where $n$ represents i.i.d noise with a mean of $\mu$ and a variance of $\sigma_n^2I$. The relation between the reference-dependent solutions in (\ref{eq_7}), i.e., $\mathbb{E}_{x|y}[\mathcal{L}(f_{\theta}(y),x)] $ and $\mathbb{E}_{\hat{x}|y}[\mathcal{L}(f_{\theta}(y),\hat{x})] $  for both $\ell_2$ and $\ell_1$  norms are discussed in the following subsections.

\subsection{Solution for $\ell_2$-norm loss function}
The proof of (\ref{eq_const_l2}) is provided below:
\begin{equation} \label{eq_l2}
\begin{split}
&\mathbb{E}_{\hat{x}|y}[\|f_{\theta}(y)-\hat{x}\|_{2}^{2}] \\ 
&= \mathbb{E}_{x,\hat{x}|y}[\|(f_{\theta}(y)-x - n)\|_{2}^{2}] \\
&= \mathbb{E}_{x,\hat{x}|y}[(f_{\theta}(y)-x - n)^T(f_{\theta}(y)-x - n)] \\
&= \mathbb{E}_{x,\hat{x}|y}[\|f_{\theta}(y)-x\|_{2}^{2} -2n^T(f_{\theta}(y)-x) + \|n\|_2^2] \\
&= \mathbb{E}_{x|y}[\|f_{\theta}(y)-x\|_{2}^{2}] -2\mathbb{E}_{x,\hat{x}|y}[n^T(f_{\theta}(y)-x)] + \mathbb{E}_{x,\hat{x}|y}[\|n\|_2^2] \\
&= \mathbb{E}_{x|y}[\|f_{\theta}(y)-x\|_{2}^{2}] -2\mathbb{E}_{x,\hat{x}|y}[n^T(f_{\theta}(y)-x) ] + d\sigma_{n}^2 + ||\mu||^2\\
&\overset{*}{=} \mathbb{E}_{x|y}[\|f_{\theta}(y)-x\|_{2}^{2}] -2(\mathbb{E}_{\hat{x}|y}[\hat{x}]-\mathbb{E}_{x|y}[x])^T\mathbb{E}_{x|y}[(f_{\theta}(y)-x)] \\
&\hspace{5mm} + d\sigma_{n}^2 + ||\mu||^2\\
&= \mathbb{E}_{x|y}[\|f_{\theta}(y)-x\|_{2}^{2}] -2\mu^T\mathbb{E}_{x|y}[f_{\theta}(y)-x] + d\sigma_{n}^2 + ||\mu||^2\\
\end{split}
\end{equation}
* Under the assumption of i.i.d noise, we can establish $\mathbb{E}_{\hat{x}|y}[\hat{x}] - \mathbb{E}_{x|y}[x]= \mathbb{E}_{x,\hat{x}|y}[n] = \mu$.

\subsection{Bounds for $\ell_1$-norm loss function}
We derive two upper bounds for $\ell_1$ loss, including (\ref{eq_const_l1}), by using the following inequality that holds for vectors $u$ and $v$ in the $p$-norm in $\mathbb{C}^n$:
\begin{equation} \label{eq_25}
\begin{split}
\mathbb{E}[\|u\|_{p}] -\mathbb{E}[\|v\|_p] \leq \mathbb{E}[\|u-v\|_{p}] 
\end{split}
\end{equation}
By setting $f_{\theta}(y)-\hat{x}$ and $f_{\theta}(y)-x$ respectively as $u$ and $v$, we can rewrite the inequality as:
\begin{equation} \label{eq_25_}
\begin{split}
\underbrace {\mathbb{E}_{(\hat{x},y)}[\|f_{\theta}(y)-\hat{x}\|_{p}] - \mathbb{E}_{(x,y)}[\|f_{\theta}(y)-x\|_{p}]}_{ 0 \leq} \leq  \mathbb{E}_{(x,\hat{x})}[\|n\|_p] \\
\end{split}
\end{equation}
Without loss of generality, we make the assumption that the training error with a corrupted reference $\hat{x}$ is greater than or equal to the training error with the clean reference $x$, leading to the non-negativity of the left-hand side of (\ref{eq_25_}).

Let $u$ be a vector in $\mathbb{C}^n$ with $1 \leq r < p$. Upon a well-known corollary of Hölder's inequality,
\begin{equation}\label{eq_35}
\|u\|_{p} \leq \|u\|_{r} \leq d^{(1/r-1/p)}\|u\|_{p},
\end{equation}
where $d$ is the dimension of $u$. By setting $p=2$ and $r=1$ in (\ref{eq_35}), we can establish a connection between the $\ell_1$ and $\ell_2$ norms as $\|u\|_{1}   \leq  \sqrt{d} \|u\|_{2}$,
which can be transformed by taking the square of each side and applying expectation rule,
\begin{equation} \label{eq_40}
\begin{split}
\mathbb{E}[\|u\|_{1}^{2}]   \leq  d \mathbb{E}[\|u\|_{2}^{2}]
\end{split}
\end{equation}
Applying Jensen's inequality, which states that $f(\mathbb{E}[x]) \leq \mathbb{E}[f(x)]$ for a convex function $f: \mathbb{R} \to \mathbb{R}$, the inequality above can be lower bounded as follows:
\begin{equation} \label{eq_41}
\begin{split}
(\mathbb{E}[\|u\|_{1}])^{2}   &\leq \mathbb{E}[\|u\|_{1}^{2}] \\  & \leq  d \mathbb{E}[\|u\|_{2}^{2}]\
\end{split}
\end{equation}
Taking the square root of both sides of (\ref{eq_41}) yields:
\begin{equation} \label{eq_42}
\begin{split}
\mathbb{E}[\|u\|_{1}]   \leq  \sqrt{d} \sqrt{\mathbb{E}[\|u\|_{2}^{2}]}
\end{split}
\end{equation}
Using the above inequality we can establish two upper bounds:
\subsubsection{Upper-bound (I)}
Considering (\ref{eq_25_}) with $p=1$ and (\ref{eq_42}),
\begin{equation}
\begin{split}
0 &\leq \mathbb{E}_{\hat{x}|y}[\|(f_{\theta}(y)-\hat{x})\|_{1}] -\mathbb{E}_{x|y}[\|(f_{\theta}(y)-x)\|_{1}] \\   &\leq   \sqrt{d} \sqrt{\mathbb{E}[\|n\|_{2}^{2}]} = 
\sqrt{d} \sqrt{d\sigma_{n}^2 + ||\mu||^2}
\end{split}
\end{equation}

\subsubsection{Upper-bound (II)}
{
Let's begin with applying inequality (\ref{eq_42}) to $f_{\theta}(y)-\hat{x}$ and $f_{\theta}(y)-x$,
\begin{subequations}
\label{eq_temp_}
\begin{align}
    \label{eq_temp_:I}
     \mathbb{E}_{\hat{x}|y}[\|f_{\theta}(y)-\hat{x}\|_{1}]   &\leq  \sqrt{d} \sqrt{\mathbb{E}_{\hat{x}|y}[\|f_{\theta}(y)-\hat{x}\|_{2}^{2}]}, \\
    \label{eq_temp_:II}
    \mathbb{E}_{x|y}[\|f_{\theta}(y)-x\|_{1}]   &\leq  \sqrt{d} \sqrt{\mathbb{E}_{x|y}[\|f_{\theta}(y)-x\|_{2}^{2}]}
\end{align}
\end{subequations}
Using (\ref{eq_temp_:I}) and (\ref{eq_temp_:II}),
\begin{equation} \label{eq_43}
\begin{split}
0 &\leq \mathbb{E}_{\hat{x}|y}[\|(f_{\theta}(y)-\hat{x})\|_{1}] -\mathbb{E}_{x|y}[\|(f_{\theta}(y)-x)\|_{1}] \\   &\leq   \sqrt{d} \bigg | \sqrt{\mathbb{E}_{\hat{x}|y}[\|(f_{\theta}(y)-\hat{x})\|_{2}^{2}]} - \sqrt{\mathbb{E}_{x|y}[\|(f_{\theta}(y)-x)\|_{2}^{2}]} \bigg |
\end{split}
\end{equation}
Inequality above can equivalently be formulated as follows:
\begin{equation} \label{eq_44}
\begin{split}
0 &\leq \mathbb{E}_{\hat{x}|y}[\|f_{\theta}(y)-\hat{x}\|_{1}] -\mathbb{E}_{x|y}[\|f_{\theta}(y)-x\|_{1}]\hspace{4mm} \\ 
&\leq{\sqrt{d} \Bigg | \frac{ \mathbb{E}_{\hat{x}|y}[\|f_{\theta}(y)-\hat{x}\|_{2}^{2}] - \mathbb{E}_{x|y}[\|f_{\theta}(y)-x\|_{2}^{2}]}{ \sqrt{\mathbb{E}_{\hat{x}|y}[\|f_{\theta}(y)-\hat{x}\|_{2}^{2}]} + \sqrt{\mathbb{E}_{x|y}[\|f_{\theta}(y)-x\|_{2}^{2}]}} \Bigg |} \\ 
&\overset{*} {=} \sqrt{d} \Bigg | \frac{ -2\mu^T\mathbb{E}_{x|y}[f_{\theta}(y)-x] + d\sigma_{n}^2 + ||\mu||^2 }{ \sqrt{\mathbb{E}_{\hat{x}|y}[\|f_{\theta}(y)-\hat{x}\|_{2}^{2}]} + \sqrt{\mathbb{E}_{x|y}[\|f_{\theta}(y)-x\|_{2}^{2}]}} \Bigg | \\
&= \frac{| -2\mu^T\mathbb{E}_{x|y}[f_{\theta}(y)-x] + d\sigma_{n}^2 + ||\mu||^2 |}{g(y,x,\hat{x}) },  \hspace{40mm}\\
\end{split}
\end{equation}
where $g(y,x,\hat{x}) = \frac{ \sqrt{\mathbb{E}_{\hat{x}|y}[\|f_{\theta}(y)-\hat{x}\|_{2}^{2}]} + \sqrt{\mathbb{E}_{x|y}[\|f_{\theta}(y)-x\|_{2}^{2}]} }{\sqrt{d}} $. * The difference between solutions for $\hat{x}$ and $x$ when loss function is $\ell_2$ norm, see (\ref{eq_l2}). Unlike upper-bound (I), (II) shows dependence on both $y$ and noise statistics.
}

\bibliographystyle{ieeetr}
\bibliography{main.bib}

\end{document}